\documentclass[conference]{IEEEtran}
\IEEEoverridecommandlockouts

\usepackage{amsmath,amssymb,amsfonts}
\usepackage{algorithmic}
\usepackage{graphicx}
\usepackage{textcomp}
\usepackage{xcolor}
\usepackage{siunitx}
\usepackage{xcolor}
\usepackage{siunitx}
\usepackage{hyperref}
\usepackage{siunitx}
\usepackage{caption}
\usepackage[list=true]{subcaption}
\usepackage{multicol}
\usepackage{multirow}
\usepackage{wrapfig}
\usepackage{gensymb}

\begin{document}

\title{PDRNN: Modular Data-driven Pedestrian Dead Reckoning on Loosely Coupled Radio- and Inertial-Signalstreams\\
{\footnotesize \thanks{This work has been carried out within the DARCII project, funding code 50NA2401, sponsored by the German Federal Ministry for Economic Affairs and Climate Action (BMWK) and supported by the German Aerospace Center (DLR), the Bundesnetzagentur (BNetzA), and the Federal Agency for Cartography and Geodesy (BKG). This work was also supported by the Bavarian Ministry for Economic Affairs, Infrastructure, Transport and Technology through the Center for Analytics Data Applications (ADA-Center) within the framework of ``BAYERN DIGITAL II'' (20-3410-2-9-8).}}}

\author{\IEEEauthorblockN{Peter Bauer\IEEEauthorrefmark{1},
    Andreas Porada\IEEEauthorrefmark{1},
    Felix Ott\IEEEauthorrefmark{1},
    Christopher Mutschler\IEEEauthorrefmark{1},
    Tobias Feigl\IEEEauthorrefmark{1}\IEEEauthorrefmark{2}
  }
  \IEEEauthorblockA{\IEEEauthorrefmark{1}Fraunhofer Institute for Integrated Circuits IIS, 90411 Nürnberg, Germany}
  \IEEEauthorblockA{\IEEEauthorrefmark{2}Friedrich-Alexander-Universität Erlangen-Nürnberg, 91058 Erlangen, Germany}
  \IEEEauthorblockA{\{andreas.porada, felix.ott, christopher.mutschler, tobias.feigl\}@iis.fraunhofer.de}
}
\maketitle
\begin{abstract}
Modern pedestrian dead reckoning (PDR) systems rely on fusing noisy and biased estimates of position, velocity, and calibrated orientation derived from loosely coupled sensors to determine the current pose of a localized object. However, discrepancies in the sampling rates of sensor-specific estimation methods and unreliable transmission pose significant challenges. And traditional methods often fail to effectively fuse multimodal sensor data during dynamic movements characterized by high accelerations, velocities, and rapidly varying orientations. 

To address these limitations, we propose a simple recurrent neural network (RNN) architecture capable of implicitly forecasting asynchronous sensor data streams from diverse estimation methods along reference trajectories. The proposed approach introduces PDRNN, a modular hybrid AI-assisted PDR system that handles each component as an independent ensemble of machine learning (ML) models to estimate both key parameter means and variances. Separate ML-based models are employed to estimate orientation, (un)directed velocity or distance from acceleration and gyroscope data, with optional absolute positioning from synchronized radio systems such as 5G for stabilization. A final fusion model combines these outputs, position, velocity, and orientation, while using uncertainty estimates to enhance system robustness. The modular design allows individual components to be updated, fine-tuned, or replaced without affecting the entire system. Experiments on dynamic sports movement data show that PDRNN achieves superior accuracy and precision compared to classic and ML-based methods, effectively avoiding error accumulation common in black-box approaches. And PDRNN offers forecast capabilities and better component control despite increased system complexity.
\end{abstract}

\section{Introduction}

Traditional PDR methods face challenges with nondeterministic dynamic motion and sensor nonlinearity due to their reliance on fixed thresholds and user-specific kinematic constraints, particularly in complex environments such as sports and virtual reality applications. They rely heavily on predefined thresholds and kinematic constraints, rendering them insufficient for non-deterministic movements and sensor nonlinearity. Consequently, they struggle to adapt to real-world scenarios, resulting in reduced accuracy and robustness. Although recent ML-based approaches such as RIDI~\cite{ferrari_ridi_2018} and RoNIN~\cite{F101_ronin} offer more sophisticated sensor data modeling and achieve higher accuracy by modeling complex movements, but treat the entire process as an black box, thereby losing control over individual processing steps and directly propagating errors to the final position. This monolithic structure sacrifices transparency and flexibility, making it challenging to control or fine-tune individual components and leading to uncontrolled error accumulation in response to sensor uncertainties or environmental changes. Consequently, classical methods tend to be overly general, employing static thresholds for generic kinematics, while modern ML-based models are often too specific, being tailored to particular individuals or scenarios. As a result, both approaches frequently fail, requiring significant effort to either recalibrate traditional methods or optimize and fine-tune black-box ML systems.

\begin{figure}[!t]
    \centering
    \includegraphics[width=1.0\linewidth]{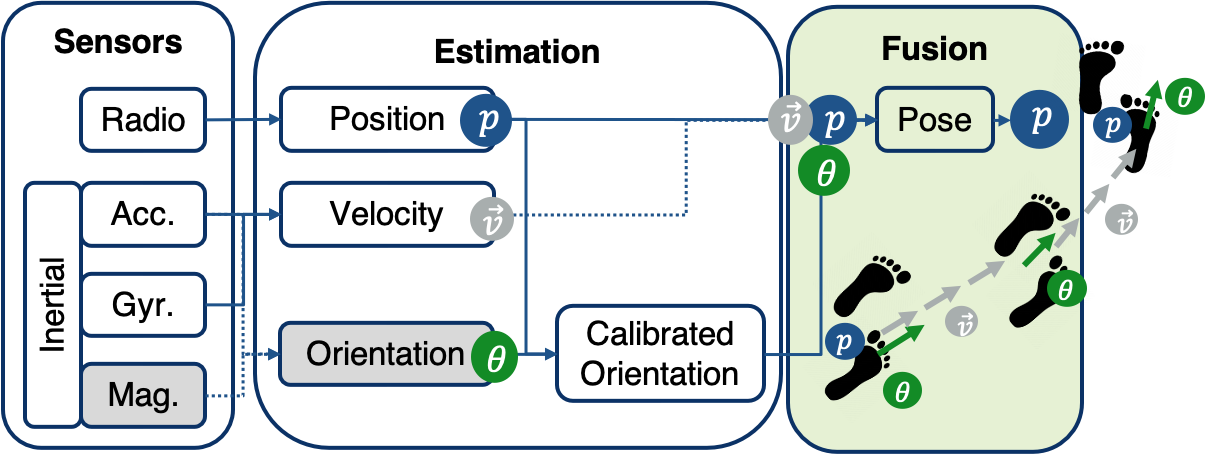}
    \caption{Overview of PDRNN: input sensor data, estimation of position, velocity, and orientation (without magnetometer), as well as pose fusion.}
    \vspace{-0.5cm}
    \label{figure_overview_introduction1}
\end{figure}

Fig.~\ref{figure_overview_introduction1} provides an overview of our approach, which addresses existing limitations by introducing a modular hybrid ML-assisted PDR system, PDRNN. PDRNN treats each component as an independent ensemble of ML-based models, which can include any type of model, to estimate both the mean and variance of key parameters. Specifically, separate ML models are employed for estimating orientation, velocity, and distance based on acceleration and gyroscope measurements, with the option to incorporate absolute positions from a time-synchronized radio frequency (RF) system, such as 5G (FR1), for initializing or stabilizing the PDR system. A final model then fuses the outputs, position, velocity, and orientation, to compute the final pose. Additionally, each (ensemble) component generates an uncertainty estimate, enhancing the robustness of the fusion process and improving error resilience. The modular design of PDRNN offers greater flexibility and control, enabling individual components to be updated, fine-tuned, or replaced without affecting the entire system, leading to more accurate and robust position estimates. Our experiments using sensor data from dynamic sports movements demonstrate that this modular ML-assisted PDR system provides more accurate and precise pose estimations than classic and ML-based methods and prevents error accumulation, a common issue in pure black-box ML approaches. Furthermore, the system's flexibility makes PDRNN highly adaptable to measurement gaps and allows for continuous improvement without necessitating a complete system overhaul.

\textbf{Contributions.} This paper contributes to human motion estimation by integrating multi-modal sensor data for accurate tracking and forecasting of pedestrian position in dynamic environments. We introduce the PDRNN architecture, combining feed-forward layers with RNN cells, enabling effective extraction of temporal features from sensor signals for accurate pose and trajectory estimation using inertial and RF measurements. We address challenges of traditional methods, such as KF and model-based PDR, which struggle with dynamic movements, sensor noise, and real-time processing. Through a data-driven approach, PDRNN shows superior robustness and accuracy, particularly in sports applications with unpredictable motion. We also demonstrate PDRNN’s ability to clean noisy sensor data and interpolate under-sampled radio signals, ensuring reliable pose estimation. PDRNN generalizes well across various movement patterns and user behaviors, enhancing its applicability in diverse contexts and laying the foundation for future advancements in forecasting. Compared to state-of-the-art methods, our approach offers higher accuracy (up to 90\%), improved resilience ($\text{CEP}_{95}$: $\text{PDR}=1.25\,m$, $\text{RoNIN}=0.46\,m$, $\text{PDRNN}=0.14\,m$) and forecasting ($\text{CEP}_{95}=0.05m$ at $1\,s$), with more control over system components, albeit at the cost of a more complex hybrid system.

\textbf{Outlook.} Section~\ref{sec:related_work} offers a comprehensive review of related literature. Section~\ref{sec:methodology} describes the proposed methodology and preprocessing pipeline. The experimental setup is detailed in Section~\ref{sec:experiments}, followed by the presentation and discussion of results in Section~\ref{sec:results}. Finally, Section~\ref{sec:summary} concludes the paper by summarizing key findings.
\section{Related Work}
\label{sec:related_work}

Section~\ref{sec:pose:probabilistische_modelle} reviews work that utilize classic, Kalman or Particle filters for dead reckoning or the fusion of inertial measurements with radio position data. Section~\ref{sec:pose:datengetriebene_modelle} presents studies that employ data-driven techniques.

\subsection{Recursive Probabilistic Methods}
\label{sec:pose:probabilistische_modelle}

To enable radio-based localization in- and outdoors, particularly in situations where radio localization is restricted or unavailable, one potential solution is dead reckoning (DR), which combines inertial with magnetometer and radio sensor data~\cite{bai_dl-rnn_2019}. DR estimates the current position based on the previous position as soon as a position change is detected~\cite{kang_smartphone-based_2018}. Consequently, the current position may include errors from both systems, such as (1) inaccuracies in step lengths and velocities due to individual body height variations, (2) radio positions affected by multipath interference, and (3) orientation errors caused by ferromagnetic interference~\cite{kang_smartphone-based_2018,alexander_stride_1984,renaudin_quaternion_2014, kang2012improved}. KFs are commonly used to merge radio and inertial sensor data. Different variants of KF are typically employed~\cite{RWMainF40,RWMainF11,RWMainF12,RWMainF0,RWMainF50} to account for the measurement noise of sensors. The measurement covariance is often optimized based on the available training data, while the movement model is defined as broadly as possible for the specific application. However, these factors result in the fusion process being highly application-specific. Even slight deviations in the sensors or the movement model lead to errors in the measurement and process covariance and significant estimation errors.

Gusenbauer et al.~\cite{gusenbauer_self-contained_2010} employ a linear KF to merge GNSS, magnetometer, and acceleration sensor data, achieving an accuracy of $\text{MAE}=4.2\,m$. Zhuang et al.~\cite{zhuang_tightly-coupled_2016} use an extended KF to merge WLAN and inertial sensor data, attaining an accuracy of $\text{MAE}=4.19\,m$ and demonstrating greater robustness against outliers compared to Li et al.~\cite{li_real-time_2015}, who achieved an accuracy of $\text{MAE}=6.2\,m$. Tao et al.~\cite{tao_lin_multiple_2015} attribute the high inaccuracies of existing methods to the distortion of magnetometer data by nearby ferromagnetic metals, which compromise the orientation estimate and, in turn, reduce positional accuracy. Sczyslo et al.~\cite{sczyslo_hybrid_2008} also use a KF to trilaterate positions derived from time of arrival (ToA) measurements of an ultra-wideband (UWB) localization system and acceleration measurements from an inertial sensor, achieving an accuracy of $\text{MAE}=0.57\,m$. The KF state vector consists of the UWB system's positions, accelerations, and the integrated speeds of the inertial sensor. Instead, Perttula et al.~\cite{perttula_distributed_2014} employ a PF to merge gyroscope and accelerometer data with ToA values from a UWB radio system, achieving different levels of accuracy depending on sensor placement: $\text{MAE}=9.9\,m$ when attached to the torso and $\text{MAE}=10.3\,m$ at the waist. This variability suggests that even a PF capable of modeling nonlinear motion is subject to significant error variance. Therefore, we utilize an optimized KF in our experiments to represent the state-of-the-art in the most accurate manner.

These probabilistic methods, that rely solely on the current and previous (Markov) states (measurements and estimations), are unable to detect data gaps or long-term data anomalies and relationships. And, the type and weighting of the fusion process are strictly predefined. Instead, our data-driven method, PDRNN, enables to learn these challenges directly and implicitly from the training data, resulting in significantly more accurate pose estimations.

\subsection{Data-driven ML-based Methods}
\label{sec:pose:datengetriebene_modelle}

While some studies estimate velocities using RNNs~\cite{kang_smartphone-based_2018} or reinforcement learning~\cite{choi_future_2018} to reconstruct distance and uncalibrated trajectories, there are few methods that leverage multimodal information~\cite{ott_fusing,ott_tro,ott_cvprw}, such as radio and inertial measurements, in data-driven approaches. Zyner et al.~\cite{zyner_long_2017} employ LSTM cells to predict future trajectories of a vehicle, using GPS positions, inertial measurement unit (IMU) orientation, and velocities from an odometry sensor to accurately reconstruct the trajectory in 90.66\% of cases. However, the reference measurements are derived from the same data, making comparisons difficult. In contrast, Yao et al.~\cite{yao_deepsense_2017} address sensor noise from signals such as GPS and IMU by utilizing a convolutional neural network (CNN) and a RNN to capture both relative movements from the IMU and global movements from the GPS system, while also modeling signal dysfunctions. Their approach achieves a mean absolute error of 4.043\,\textit{m} ($\text{SD}=0.524\,m$), compared to the mean absolute error of 6.065\,\textit{m} ($\text{SD}=0.565\,m$) observed in traditional sensor data fusion methods. As these preliminary studies focus on vehicle movements, it remains unclear whether and how their concepts apply to human motion. 

The central concept of our paper aligns with the approaches of Zyner et al.~\cite{zyner_long_2017} and Yao et al.~\cite{yao_deepsense_2017}, but extends their methodologies to narrowband (5G / FR1) and UWB radio localization as well as 6DoF inertial sensors. Unlike previous work, we conduct a comprehensive architecture and parameter study to identify the optimal light-weight ML framework for these specific domains. Additionally, we explore the transient behavior in highly dynamic scenarios and examine the impact of input data stream sequence length on the accuracy and robustness of data-driven PDR. Our approach builds upon these existing methodologies by introducing a modular and hybrid PDR system that integrates radio and inertial sensor data through independent ensembles of ML models. Essentially, the literature review suggests that our paper represents one of the first publicly known studies to enable accurate and reliable detection, tracking, and fusion of human motion using data-driven methods in highly dynamic scenarios, characterized by sudden changes in speed and direction, noisy inertial measurements from loosely placed sensors, and radio signals in multipath environments.
\section{Methodology}
\label{sec:methodology}

Section~\ref{label_problem_description} first outlines the problem, addressing data gaps, asynchronous data streams, and nondeterministic calibration. Subsequently, Section~\ref{label_method_section} presents our proposed methodology. Section~\ref{sec:pose:methods:sequential} presents classical approaches and Section~\ref{sec:pose:methods:frfnn} presents our ML-driven method.

\subsection{Problem Description}
\label{label_problem_description}

Dead reckoning forms the foundation for many modern navigation applications, including pedestrian, vehicle, and airplane navigation. If the initial position $p_0$ of an object is known, the subsequent absolute poses $p_1, p_2, \ldots, p_n $ can be determined using orientation, velocity $v$, acceleration $acc$, or distance covered $d$. The future absolute pose is predicted based on previous poses $p$~\cite{bohm_handbuch_1978}. Environmental factors, such as sensor interference, temporary signal loss, attenuation, scattering, diffraction, refraction, reflection, drift, and sensor noise, lead to deviations in course or position, making fusion localization necessary for fail-safe operation~\cite{jiang_mixed_2019}. IMUs correct localization system measurements, and vice versa~\cite{li_uwbpdr_2018}, with IMUs regularly recalibrated using complementary localization technologies~\cite{li_uwbpdr_2018}. As radio-based localization is also susceptible to signal interference, inertial measurements help sustain dead reckoning, particularly during multipath propagation~\cite{li_uwbpdr_2018}.

\textbf{Data Gaps and Asynchronous Data Streams.} Existing fusion methods face several challenges. Sensors produce measurements at varying data rates that cannot be processed by predefined filters, as model-driven approaches require time synchronization between different sensors and uninterrupted data streams. As most modern radio and inertial systems are loosely coupled, the radio position is determined by an external system and must be sent back to the object to be localized, or inertial measurements must be transmitted to the localization environment along with the emitted radio signals. Both variants may introduce nondeterministic delays that hinder real-time localization, as such delays can only be approximated by a rigid model. In the worst case, data streams become asynchronous, and individual data points are lost.

\textbf{Nondeterministic Calibration.} It remains unclear when and to what extent a current position or orientation should optimally contribute to correction and calibration to rectify accumulated errors in an inertial measurement system or incorrect radio positions, thereby reconstructing poses and trajectories accurately. Correcting the current pose estimate is only effective if the correction information (such as positions or velocities) are free of errors or more precise than the current pose. Classic, manually-designed methods often incorrectly determine these relationships. For instance, errors in the inertial measurement system depend on application-specific movement dynamics, where changes in speed and direction may lead to larger errors, while errors in the radio system are influenced by environment-specific signal propagation properties. Furthermore, constant calibration of the velocity estimate with a position does not always make sense, as the radio position typically introduces larger errors than the velocity estimate over short time intervals. Thus, determining when to calibrate with a position to optimize pose accuracy and robustness is not straightforward. It is also unclear how to best weight and update the position and orientation relative to the current velocity or step length estimate for calibration or correction. A simplistic adjustment of the inertial sensor system’s current position estimate to the current radio position may result in significant position jumps.

\begin{figure}[!t]
    \centering
    \includegraphics[width=1.0\linewidth]{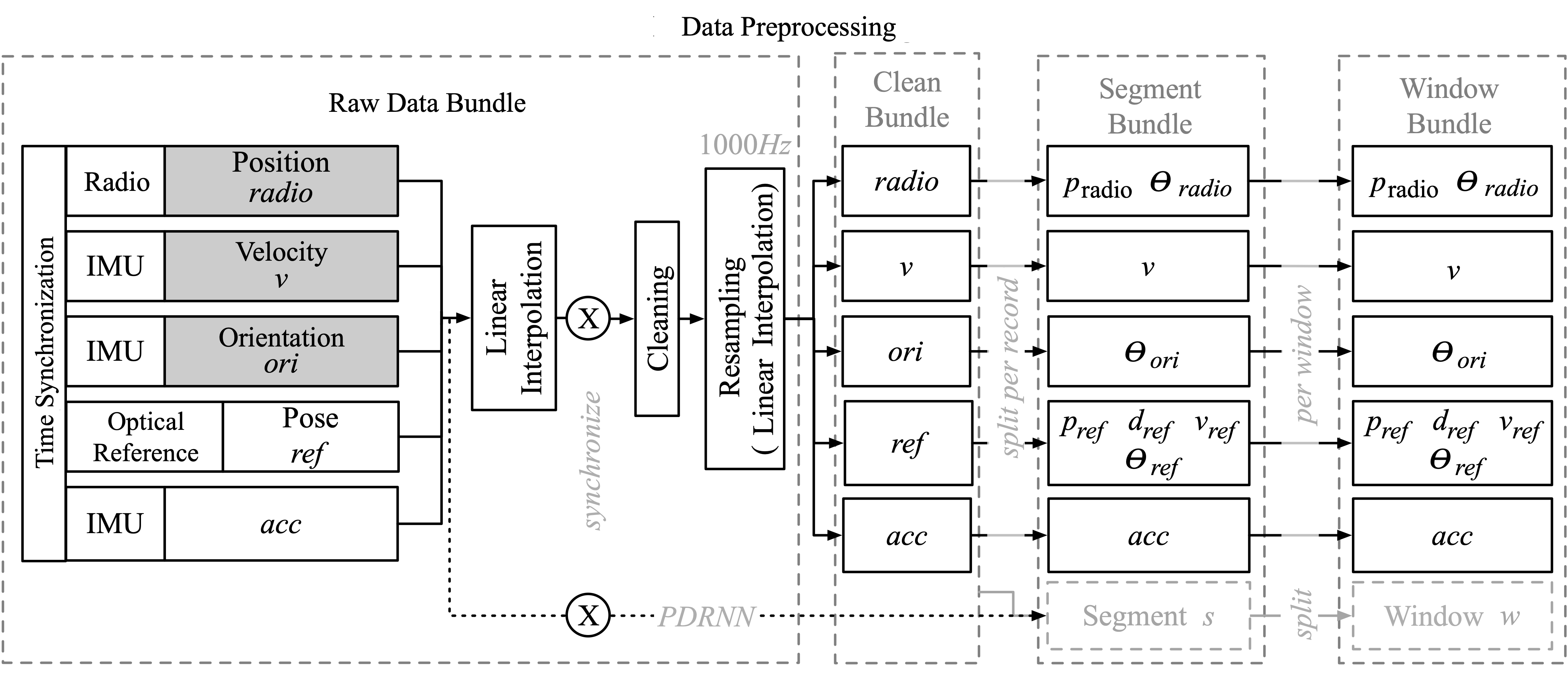}
    \caption{Data preprocessing pipeline. Note that PDRNN implicitly fuses raw information streams (dashed line at the bottom).}
    \vspace{-0.5cm}
    \label{figure_overview_introduction}
\end{figure}

\subsection{Method Overview}
\label{label_method_section}

\textbf{Proposed Solution.} The core concept of our approach for precise and robust pose estimation is the integration of low-frequency data-driven position tracking (using e.g., 5G / FR1 radio signals) with high-frequency data-driven relative velocity estimation (based on acceleration and gyroscope measurements from a smartphone), as well as long-term stable, low-frequency data-driven yaw orientation estimation (also derived from acceleration and gyroscope measurements). We propose a data-driven methodology, PDRNN, capable of processing sensor data with varying data rates, non-synchronized measurements, and variable time delays, while implicitly determining the optimal usage of absolute radio positions, (un)directed relative velocities, and (optionally) absolute orientations to predict an accurate absolute pose. The key idea is to combine input streams, such as position, velocity, and orientation, and map these onto reference data streams, such as (future) reference positions, using a data-driven model. Our method leverages both Feedforward (FF) and LSTM cells to reduce capacity requirements, capture both short- and long-term spatio-temporal correlations in the data, and implicitly identify and compensate for data gaps and other artifacts. The data-driven approach autonomously learns how and when correction information should be integrated into the fusion process to estimate an optimal pose. Our modular PDRNN utilizes ensembles of ML models at each stage of the pipeline that are most appropriate for each task.

\textbf{Pipeline.} PDRNN learns to map specific local sensor data streams to corresponding tasks, including orientation, velocity / distance, position, and the final pose ({directed distance in absolute world coordinates}). The model outputs future 2D body poses, providing forecasts of human movement trajectories up to 2\,\textit{s} ahead. It integrates radio and inertial sensors in a modular, data-driven pipeline to estimate position, velocity, and orientation under dynamic movement conditions. Especially, when the individual sensor data streams are loosely coupled and aggregated in real-time, without the need for strict or reference-based time synchronization. As a result, PDRNN is designed to handle asynchronous or missing information. Radio position measurements (at 10\,Hz) experience a motion-to-photon delay of $98-244\,ms$, while inertial sensors (at 100\,Hz) exhibit delays ranging from $5-13\,ms$.

\textbf{Data Preprocessing.} The data processing is performed for the individual input variables, including position~\cite{feigl_recurrent_2018}, velocity~\cite{feigl_bidirectional_2019}, and orientation~\cite{feigl_head--body-pose_2018}. For a fair comparison of PDRNN to state-of-the-art methods, we have to preprocess the data. However, note that PDRNN is running on direct data streams without further preprocessing, see the dashed line in the bottom of Fig.~\ref{figure_overview_introduction}. For all other methods, data preprocessing involves time-synchronizing the input data streams (radio position $p_\text{radio}$ from the radio system $radio$, velocity $v$, orientation $\theta_\text{ori}$, reference pose $ref$, and acceleration $acc$), cleaning them, and preparing them through resampling and interpolation before bundling the cleaned data streams. These clean data bundles are further processed into segment ($s$) and window ($w$) bundles, as described by Feigl et al.~\cite{feigl_bidirectional_2019}. Each bundle variant contains radio positions $p_\text{radio}$, radio orientation $\theta_\text{radio}$ (representing the direction of movement between two consecutive $p_\text{radio}$), velocity estimate $v$, orientation estimate $\theta_\text{ori}$, reference pose $ref$ (with $p_\text{ref}$, $d_\text{ref}$, $v_\text{ref}$, and $\theta_\text{ref}$), and raw 3D acceleration data $acc$. The orientation $\theta_\text{ori}$ is estimated based on the algorithm of Lavalle et al.~\cite{lavalle2014head}, and the current orientation is calibrated for all window bundles, which are classified using an ensemble of cubic support vector machines~\cite{feigl_head--body-pose_2018}. The data streams $p_\text{radio}$, $v$, $\theta_\text{ori}$, and (optionally) $acc$ are merged and evaluated against the reference pose $ref$. Segments $s$ represent the clean, preprocessed data from each recording, such as that of an athlete with radio and inertial sensors. The windows $w$ represent input sequences for pose estimation, which slide along the segment $s$. To process in real-time, time synchronization occurs just before linear interpolation, where data are merged as they arrive and subsequently processed into segments or windows.

\textbf{(Initial) Position(s).} The pipeline initiates with the real-time aggregation of sensor data, where radio signals are captured by stationary antennas to estimate position using the Uplink TDoA method. This method synchronizes the signal arrival times to determine the transmitter's location through a tiny and simple sequence-to-sequence LSTM-based learning algorithm~\cite{feigl_position}, ensuring robust positioning, even during high-speed activities with sudden stops and directional changes.

\textbf{(Relative) Velocities.} Concurrently, the modular ML-assisted pipeline estimates velocities by utilizing sliding windows of acceleration and gyroscope data from inertial sensors, focusing on the relative magnitudes of the sensor's motion independently of orientation. A novel, compact, and efficient sequence-to-one ResNet-based method~\cite{resnet1d,feigl_velocity2} ensures high performance even under dynamic conditions.

\textbf{(Calibrated) Orientation.} Simultaneously, a Madgwick filter~\cite{madgwick2010efficient} is employed to estimate the orientation by integrating acceleration, rotation rate, and magnetic field data. To mitigate potential magnetic field interference, an additional calibration step adjusts the estimated orientation based on body movement direction classification and walking direction~\cite{feigl_orientation1,feigl_orientation2}. This step corrects any misalignment between body and head orientation by comparing consecutive positions and estimating a calibrated direction, thereby ensuring accurate alignment with the positioning coordinate system.

\textbf{(Final) Pose/Position.} The pipeline integrates position, velocity, and calibrated orientation data to generate a continuous pose estimate. As position updates from radio signals are typically less frequent (below 10\,Hz), the system interpolates the position between updates using velocity and orientation, thereby refining the position estimate until the next radio position-based calibration. PDRNN incorporates a robust fusion method that ensures high accuracy in dynamic environments, such as sports, where rapid changes in velocity, orientation, and acceleration are common. The variable delays in the incoming sensor data streams are compensated for by PDRNN’s capability to forecast positions beyond $250\,ms$ and implicitly perform interpolation.

\subsection{Classical Model-based Approach}
\label{sec:pose:methods:sequential}

To reconstruct poses and trajectory, we use $p_0$ from the $radio$-system~\cite{feigl_position} as the starting position and we transform $\theta$ and $\rho$, where $\rho$ = distance derived from $v$, $\vec{v}=1\frac{m}{s} \rightarrow \vec{d}=1\,m$, with $dt=1.0\,s$ and $\theta \in \{\theta_\text{ori}, \theta_\text{radio}, \theta_\text{ref}\}$ per window):
\begin{equation}
    {p(x)} = p_0(x)+ \rho\cdot\cos(\theta),
\end{equation}
\begin{equation}
    {p(y)} = p_0(y) +\rho\cdot\sin(\theta).
\end{equation}
If the orientation $\theta_\text{ori}$ is used for the calibration, $\theta$ is corrected accordingly $\theta = \theta_{ori}$ and $p(x)$ and $p(y)$ are adjusted by the new $\theta$. If the position is calibrated with a current radio position $p_\text{radio}$, $p(x)$ and $p(y)$ are set accordingly: ${p(x)} = p_\text{radio}(x)$ and ${p(y)} = p_\text{radio}(y)$. For a fair comparison we employ the following models of Feigl et al.~\cite{feigl_velocity1,feigl_velocity2}: an optimized KF and PDR, ML-GP, RoNIN, and C/RNN. Note that these models only estimate (un)directed velocities. 

\subsection{Machine Learning-based Approach}
\label{sec:pose:methods:frfnn}

\begin{figure}[t!]
    \centering
    \includegraphics[width=1.0\linewidth]{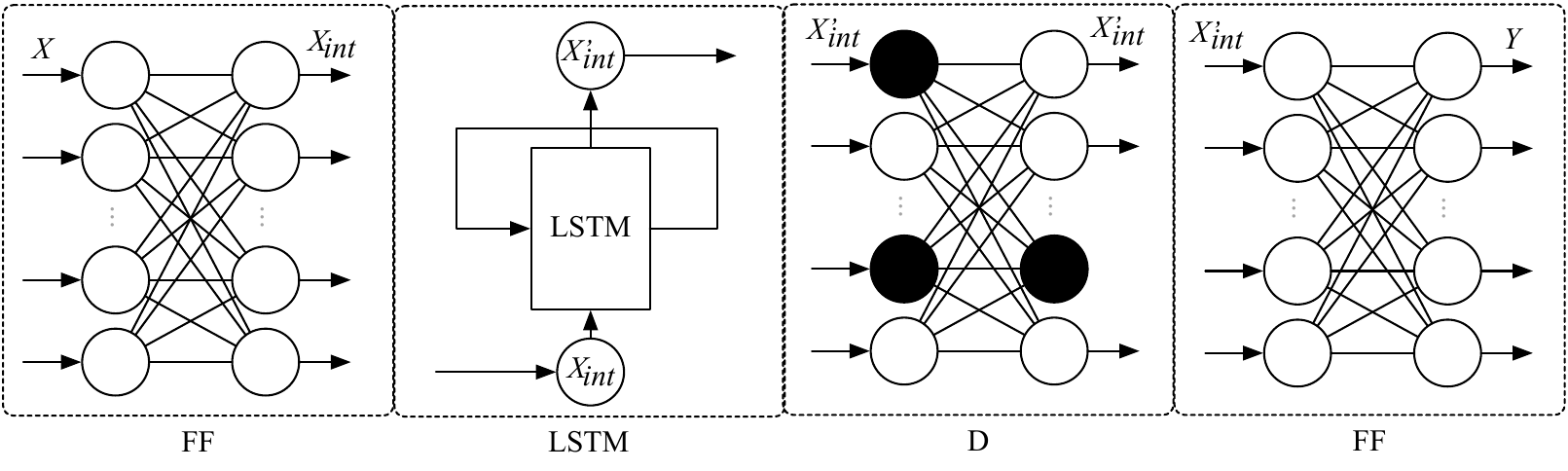}
    \caption{PDRNN architecture to fuse multimodal sensor signals. FF layers process the input and output of the LSTM layer. A dropout layer (D) after the LSTM layer reduces the possibility of overfitting and allows uncertainty estimation.}
    \label{fig:pose:methods:frfnn}
\end{figure}

To derive an optimal variant of PDRNN that is tiny, fast, and energy efficient, the following recurrent cell architectures were evaluated: vanilla RNN~\cite{elman1990finding}, IRNN~\cite{le2015simple}, LSTM~\cite{hochreiter1997long}, GRU~\cite{cho2014learning}, as well as cell concepts of Jozefowicz et al.~\cite{jozefowicz2015empirical} and Greff et al.~\cite{greff2017search}. In addition to the cell architecture, the structure that connects and incorporates the cells is critical for optimizing the model architecture. While standard RNN cells have been successfully applied to a variety of tasks, ranging from artificial addition tasks to music generation~\cite{hochreiter1997long,jozefowicz2015empirical,le2015simple}, many alternative networks have been proposed that use recurrent cells for more complex designs. These networks include variations such as deep RNNs~\cite{hermans2013training,pascanu2013deep} and specialized architectures such as bidirectional RNNs~\cite{schuster1997bidirectional}, encoder-decoder networks~\cite{cho2014learning,sutskever2014sequence}, and attention-based models~\cite{bahdanau2014neural}. We adapted these existing methods to pose estimation tasks. Inspired by Pascanu et al.~\cite{pascanu2013deep,pascanu_how_2014}, we incorporate deep FF networks alongside stacked RNNs for the transition functions, to process both input and output, and FF layers are placed directly after the input layer and before the output layer. This approach enables the network to perform more complex computations between timesteps, enhancing accuracy. In contrast to Pascanu et al.~\cite{pascanu2013deep}, we do not use depth connection functions, as they promote vanishing and exploding gradients, which complicate training. The FF and RNN layers are stacked, with the outputs from one layer fed to the next in each processing step. Particularly for tasks requiring extensive processing of input data, deeply stacked RNNs have outperformed single-layer RNNs~\cite{my_5,sutskever2014sequence}.

\begin{figure}[!t]
    \centering
    \includegraphics[width=1.0\linewidth]{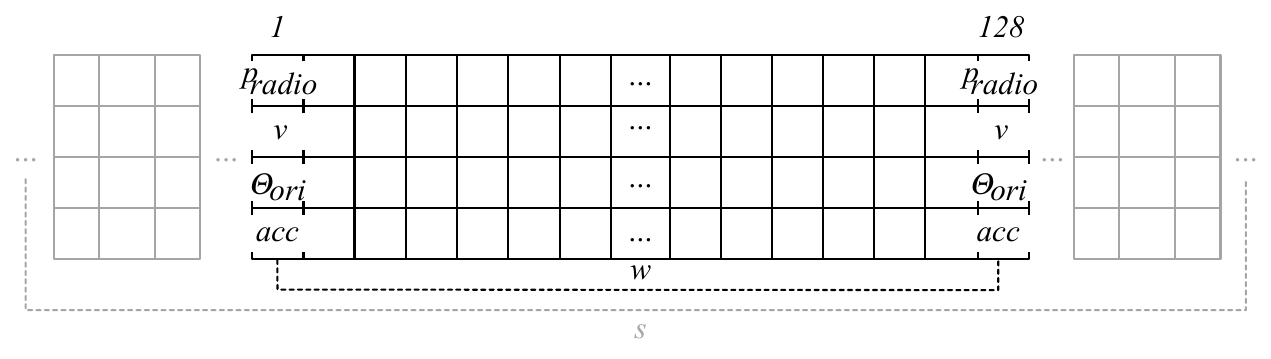}
    \caption{Input data ($X$) of the data-driven pose estimator. A window $w$ with a length of 128 timesteps slides over the segment $s$ with the signal vectors.}
    \label{fig:pose:methods:input}
\end{figure}

The core concept of the PDRNN architecture is to utilize upstream and downstream FF layers to transform the input signal $X$ (that may include radio position $p_\text{radio}$, velocity $v$, acceleration $acc$, and orientation $\theta_\text{ori}$) from its high-dimensional form into an optimal internal dimension $X_\text{int}$. The final architecture is depicted in Fig.~\ref{fig:pose:methods:frfnn}. This internal representation is then processed by the RNN architecture, that optimally handles the representation and forwards it to the downstream FF layer. The downstream FF layer subsequently transforms the RNN’s internal representation $X^\prime_\text{int}$ into the target representation (pose) $Y$ with a different dimensionality. The multimodal input domain, consisting of $p_\text{radio}$, $v$, $acc$, and $\theta_\text{ori}$, is thus mapped to a target domain (pose) through an RNN architecture that maximizes accuracy by combining upstream FF layers before LSTM cells and downstream FF layers. A dropout layer placed between the LSTM cells and the downstream FF layer helps mitigate overfitting and enables the uncertainty of the method to be assessed. Fig.~\ref{fig:pose:methods:input} illustrates the structure of an exemplary input sequence $X$. A sliding window $w$ moves over the segment $s$ data. Each window contains multidimensional vectors representing the input variables over, e.g., 128 timesteps.
\section{Experiments}
\label{sec:experiments}

Section~\ref{label_exp_datasets} describes the dataset. Section~\ref{sec:pose:experiment:params} introduces the configuration of PDRNN.  Section~\ref{sec:pose:experiment:paramsother} presents the configuration of all other (state-of-the-art) methods.

\subsection{Dataset}
\label{label_exp_datasets}

\begin{figure}[!t]
    \centering
	\begin{minipage}[t]{0.235\linewidth}
        \centering
    	\includegraphics[width=1.0\linewidth]{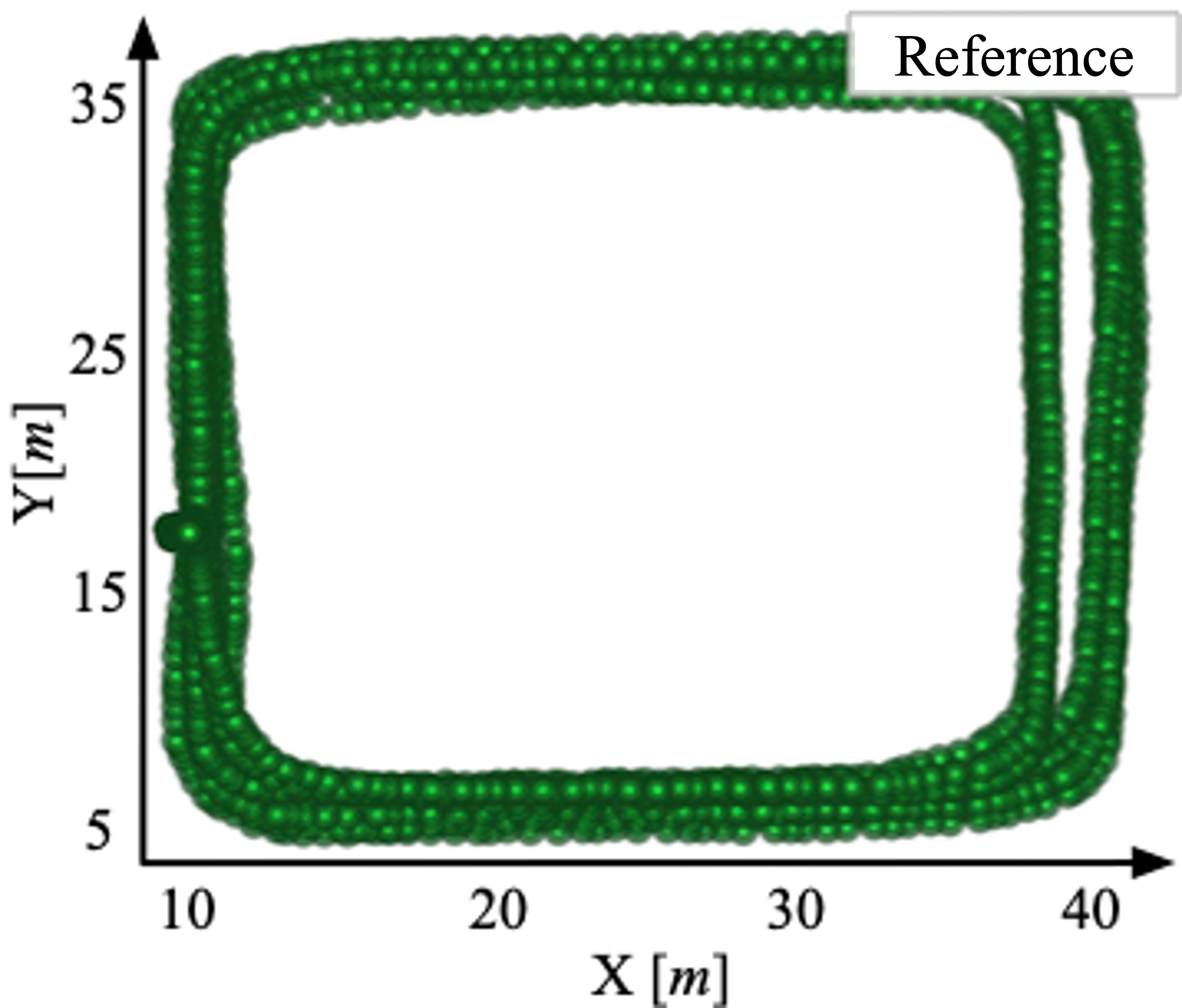}
        \subcaption{Walking.}
    \end{minipage}
    \hfill
	\begin{minipage}[t]{0.235\linewidth}
        \centering
    	\includegraphics[width=1.0\linewidth]{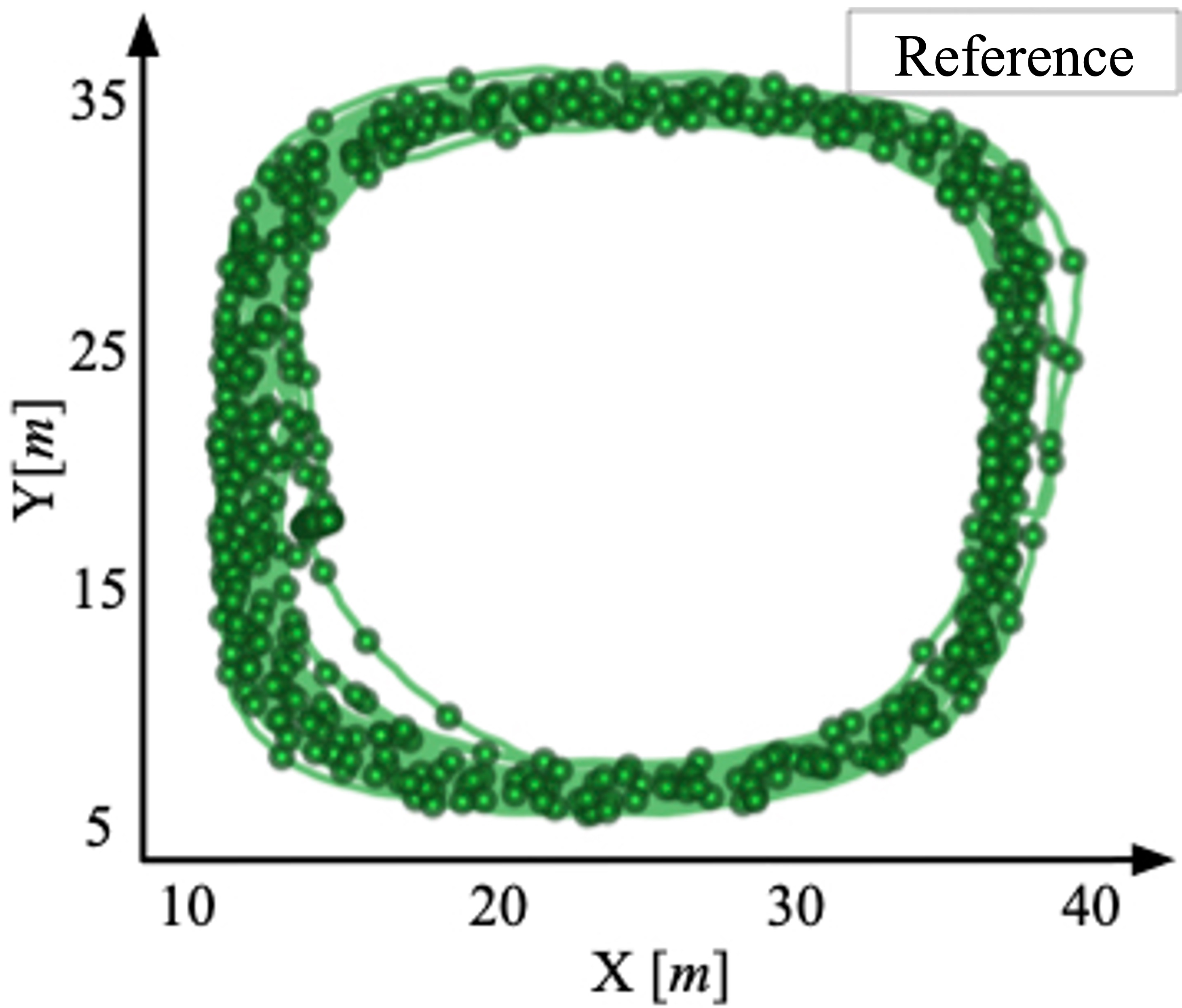}
        \subcaption{Jogging.}
    \end{minipage}
    \hfill
	\begin{minipage}[t]{0.235\linewidth}
        \centering
    	\includegraphics[width=1.0\linewidth]{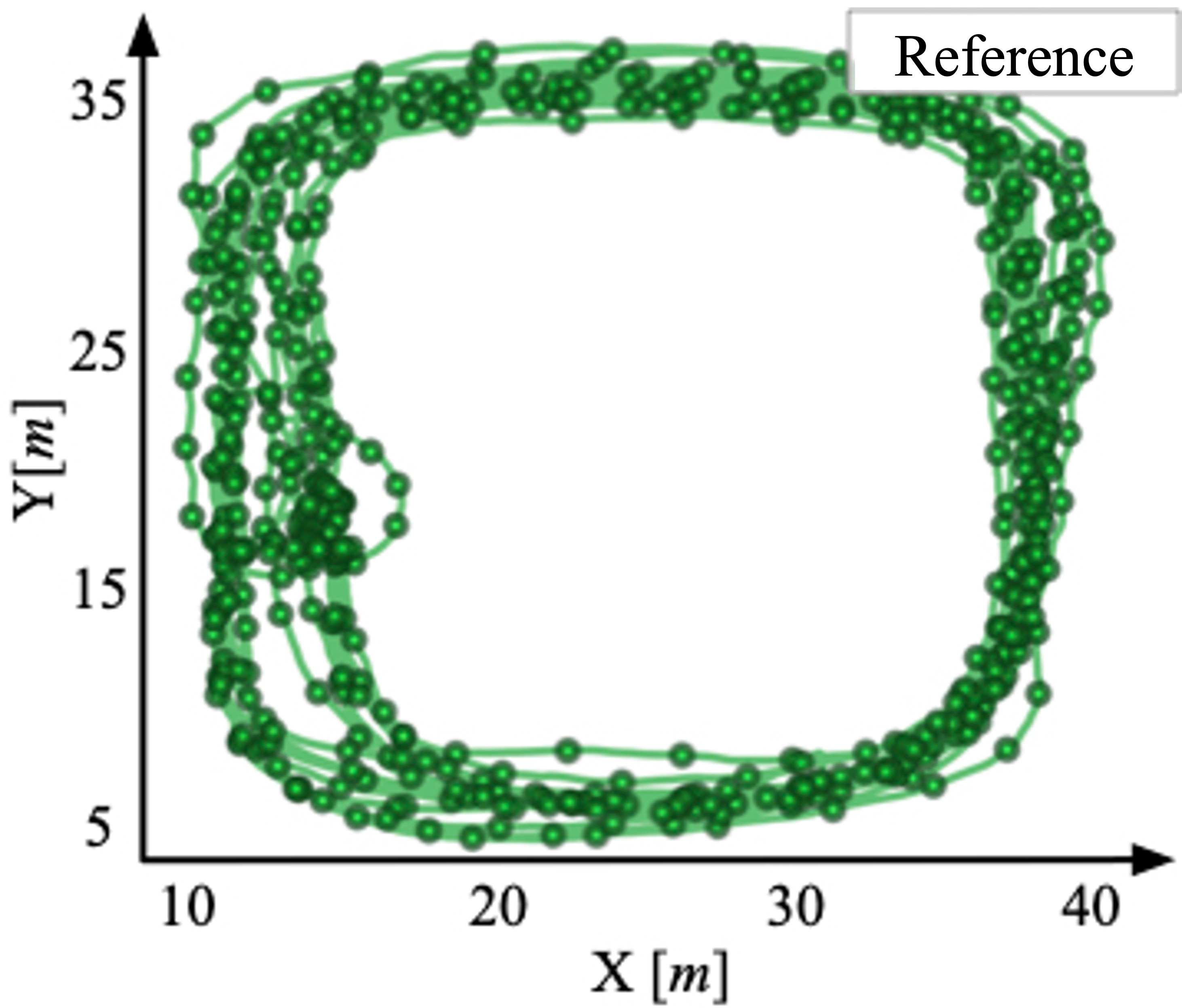}
        \subcaption{Running.}
    \end{minipage}
    \hfill
	\begin{minipage}[t]{0.235\linewidth}
        \centering
    	\includegraphics[width=1.0\linewidth]{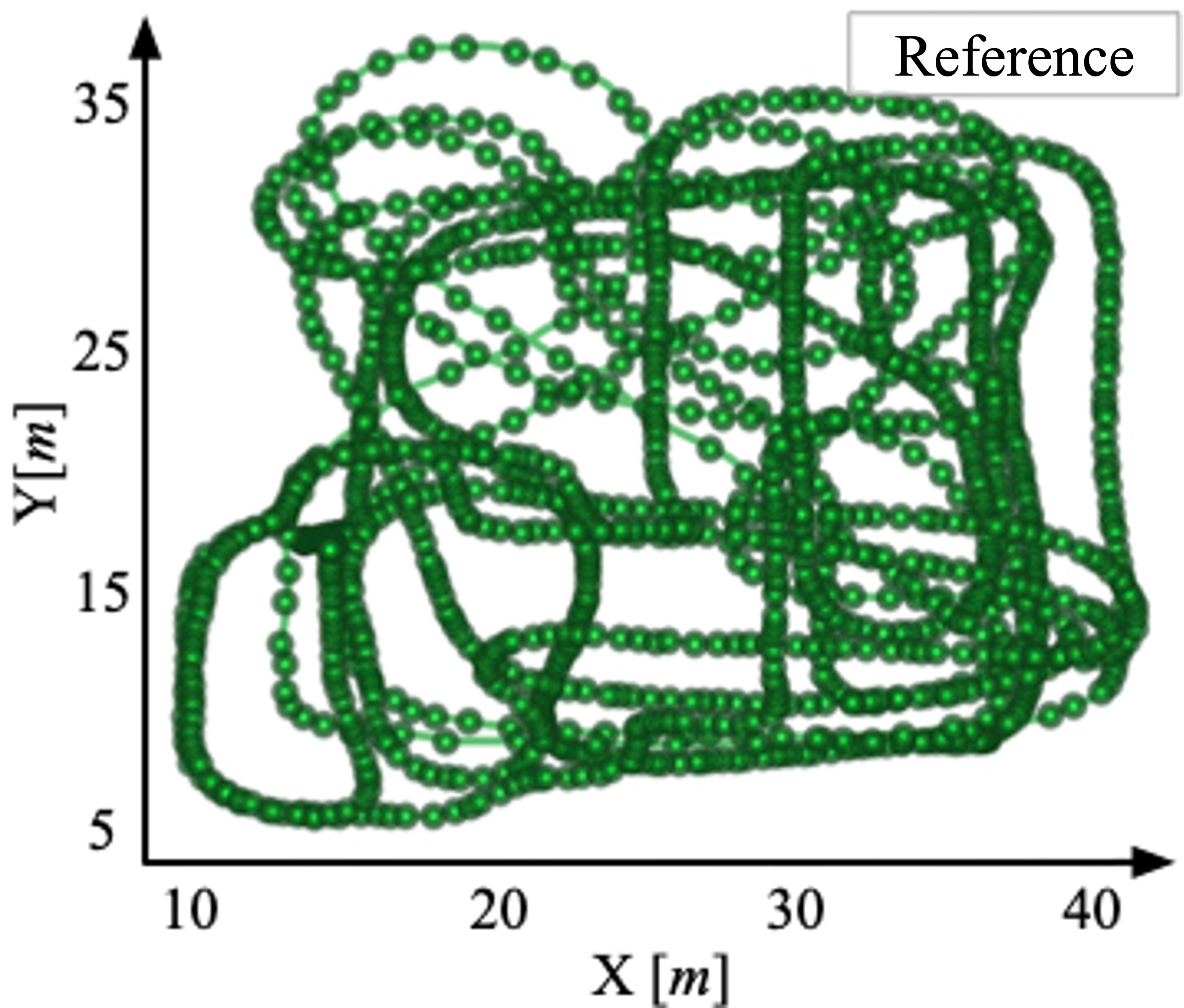}
        \subcaption{Random.}
    \end{minipage}
	\begin{minipage}[t]{0.235\linewidth}
        \centering
    	\includegraphics[width=1.0\linewidth]{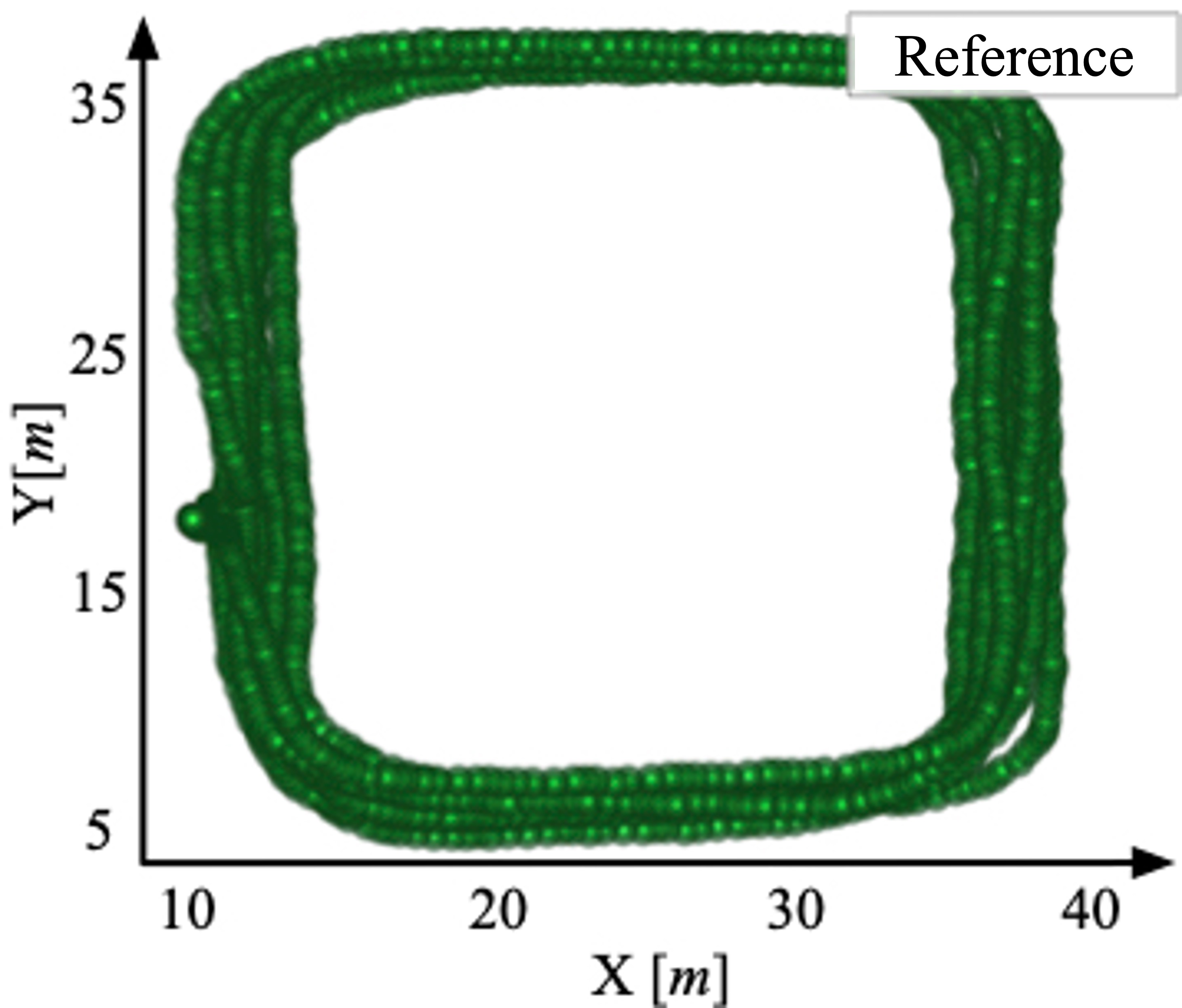}
        \subcaption{Walking.}
    \end{minipage}
    \hfill
	\begin{minipage}[t]{0.235\linewidth}
        \centering
    	\includegraphics[width=1.0\linewidth]{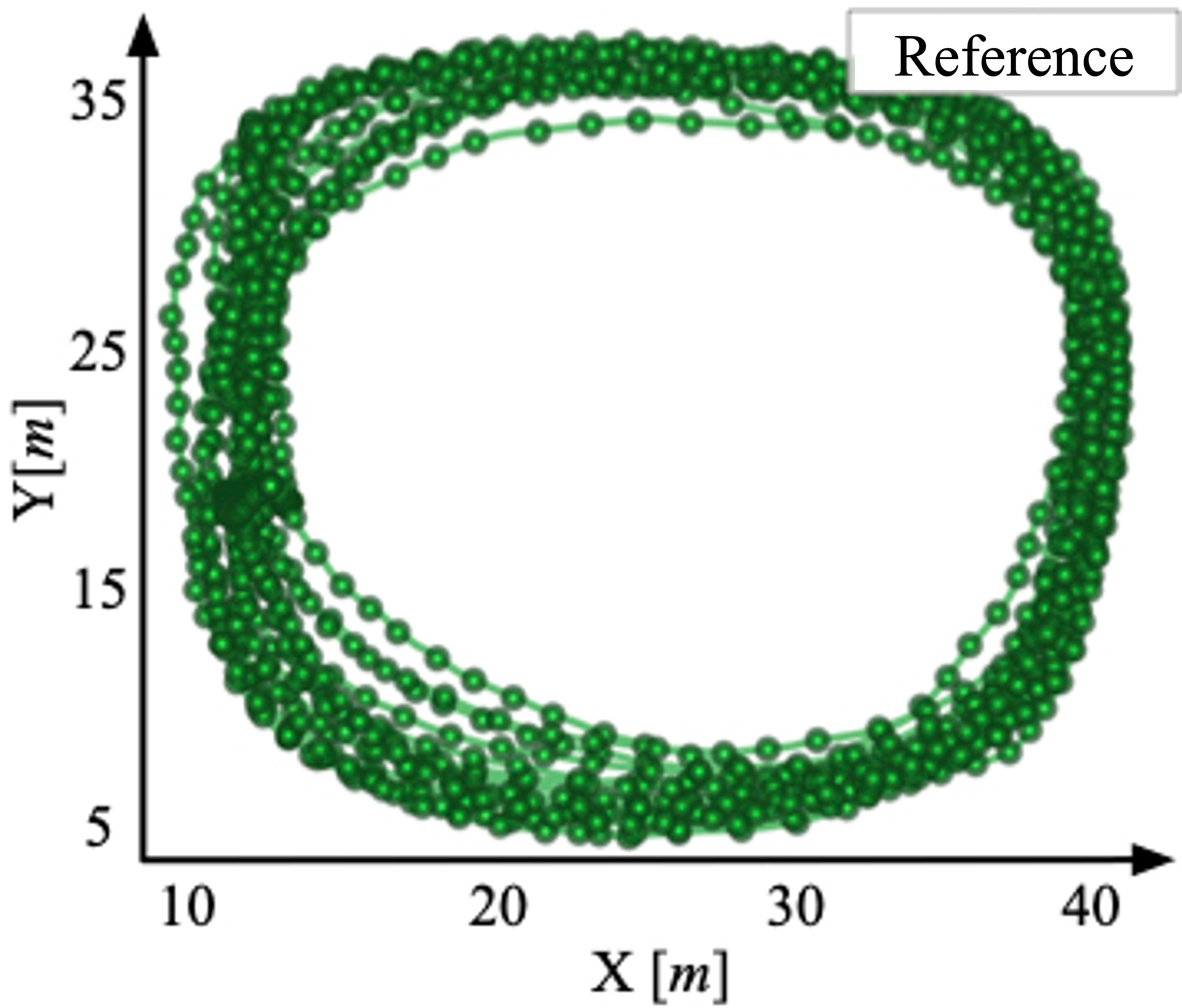}
        \subcaption{Jogging.}
    \end{minipage}
    \hfill
	\begin{minipage}[t]{0.235\linewidth}
        \centering
    	\includegraphics[width=1.0\linewidth]{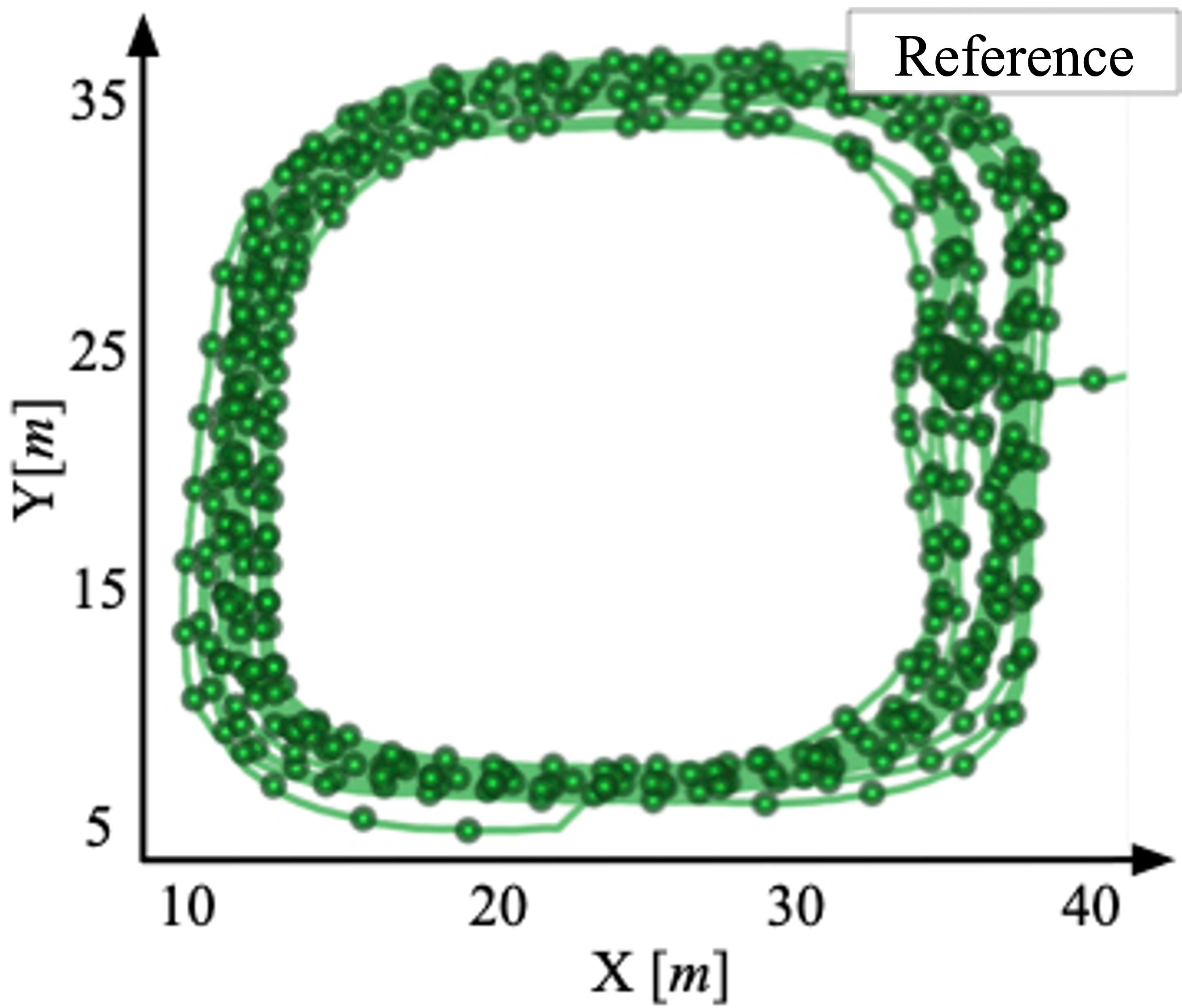}
        \subcaption{Running.}
    \end{minipage}
    \hfill
	\begin{minipage}[t]{0.235\linewidth}
        \centering
    	\includegraphics[width=1.0\linewidth]{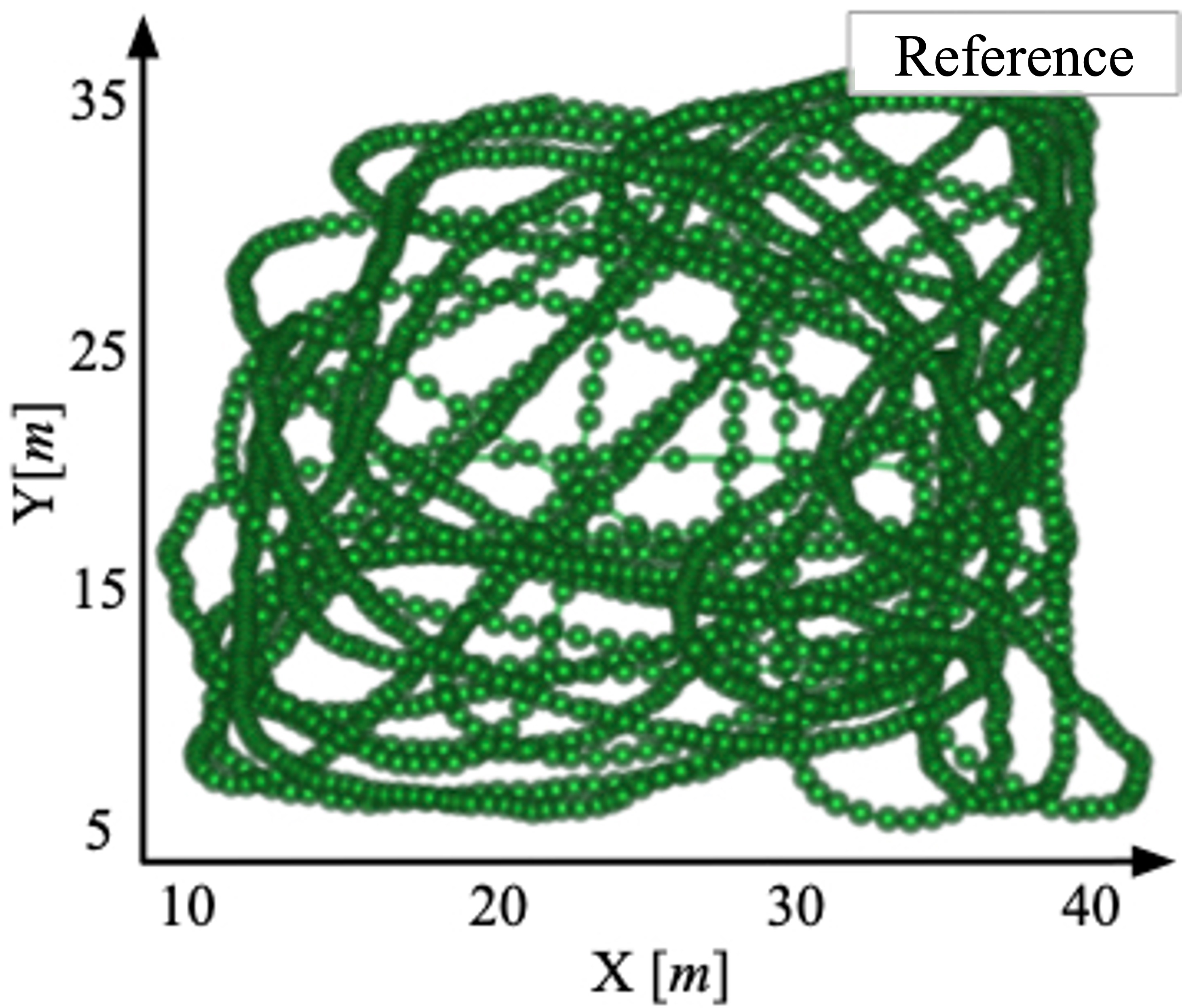}
        \subcaption{Random.}
    \end{minipage}
    \caption{Exemplary reference trajectories for two segments $s$.}
	\label{fig:pose:dataset-examples}
\end{figure}

The dataset, proposed by Feigl et al.~\cite{feigl2020rnn}, consists of motion data collected from 23 participants (17 male, 6 female; average age 26.7 years; height range: $1.46\,m$ to $1.87\,m$), using an optical reference system, Google smartphone IMUs, and 5G radio signaling. The participants performed four activities: walking, jogging, running, and random movements. Each activity lasted approximately 7.5 minutes, beginning with a 1-minute stationary period for sensor calibration and initial error estimation. Data were collected with two sensors per participant, resulting in a total of about 59 hours of recordings and covering approximately $107\,km$. The dataset contains roughly 2,674,688 overlapping windows, enabling a comprehensive analysis of motion patterns and velocity. Exemplary trajectories from the dataset are shown in Figure~\ref{fig:pose:dataset-examples}.

We found that sliding windows of size 128\,Hz with a 50\% overlap ($N_w \cdot 0.5 = 64$) provide excellent features with low computational costs, resulting in the highest accuracy for velocity estimation. This configuration, with a window size of $1.28\,s$, effectively captures long-term relationships in human movement. Additionally, we explore sliding windows in the range from $0.5\,s$ to $30\,s$, with overlaps ranging from one timestep to 100\% ($N_w \cdot 1.0 = 128$).

\subsection{Parameterization of PDRNN}
\label{sec:pose:experiment:params}

To identify the optimal architecture and parameters for the PDRNN method, a grid search was performed. \footnote{PDRNN parameters (best in bold): Solver: SGD, \textbf{Adam}, Adadelta, RMSProp; $\beta_1 = \textbf{0.01}$; $\beta_2 = 0.01$; Momentum: \textbf{0.9}; FF, CNN, TCN, RNN, IRNN, GRU, LSTM, BLSTM layers: [1, 2, \textbf{3}, 4, 8, 16]; Cells per layer: \textbf{120} [1:1:50, 50:10:500]; Initial learning rate: \textbf{0.001} [1.0:0.1:0.00001]; LR reduction: Learning rate reduces every $n$ periods, where $n \in [0, 10, 50, 100]$ epochs; LR reduction rate per period: \textbf{0.5}; Batch size: [128, 256, 512, \textbf{1,024}, 2,048]; Regularization: [$L_1$, \textbf{$L_2$}, Huber loss, Log-cosh loss]; Parameter: [100, 1,000, 10,000, \textbf{100,000}, 1,000,000]; Shuffle: [no, \textbf{per epoch}]; Gradient clipping: max (input); Dropout layer: [before, between, \textbf{after} RNN layers]; Dropout rate: $\textbf{0.5} \in [0.1, 0.7]$; Each combination of $f_s$ (50, \textbf{100}, 200, 400) and $N_w$ (64, \textbf{128}, 256, 512).}
Each parameter set was trained for a maximum of 100 epochs, with early stopping. The resulting best combination of FF+LSTM+FF (with 120 LSTM cells) provides the most accurate pose estimates with short inference times.

\subsection{Parameterization of Other Methods} 
\label{sec:pose:experiment:paramsother}

\textbf{Position Estimation.} We employ the same parametrization and architecture of the radio-based positioning model of Feigl et al.~\cite{pos_tof__rnn_1_faucris.202456629}. Their position estimator yields radio positions with an error of MAE=0.1731 $m$ (SD = 0.031 $m$) for V2 and an error of MAE=0.1477 $m$ (SD = 0.028 $m$) for V3.

\textbf{Orientation Estimation and Calibration.} We employ the implementation of the orientation estimation of Lavalle et al.~\cite{lavalle2014head}. Their orientation estimation yields orientations with an error of MAE=1.87\textdegree (SD = 0.96\textdegree) for V2 and an error of MAE=1.41\textdegree (SD = 0.78\textdegree) for V3. To calibrate the orientation, we employ the same parametrization and architecture of Feigl et al.~\cite{feigl_orientation1, feigl_orientation2}. The calibration of the orientation using radio position estimates (see above) based on Feigl et al.~\cite{pos_tof__rnn_1_faucris.202456629} with an ensemble of 100 cubic support vector machines, yields correct orientations for V2 in 96.7 of 100 cases (SD = 4.01\textdegree) and for V3 in 98.3 of 100 cases (SD = 3.89\textdegree).

\textbf{Velocity Estimation.} In total, we evaluated four different velocity estimation methods: an optimized PDR, ML-GP, RoNIN, and C/RNN. Details of the parametrization and architecture of the models are listed by Feigl et al.~\cite{feigl_velocity1}. The errors of these velocity estimation methods are given in Section~\ref{sec:results} in Table~\ref{table:pose:results:sequential:results_accuracy_trajectory}.

\textbf{Pose Estimation.} A linear KF model~\cite{coskun2017long} from the state-of-the-art is employed. Under the assumption of drift-free signals, these models are considered optimal state estimators. This Bayesian model characterizes the motion transition function (motion model) and the development of measurement and process noise as linear functions, influenced by Gaussian noise~\cite{feigl2020rnn}. To optimize the KF for the respective training data, it is initially configured with a starting state of $x_0 = 0$, covariance $P = 1$, process noise $Q = 0.1$, measurement noise $R = \sigma = 0.1$, and the transition function defined by a constant velocity. Given the availability of empirical knowledge about the data (i.e., the training dataset), the KF is tailored to optimally suit the specific training data. This configuration ensures that the KF model becomes an optimal estimator for the respective training set. For this purpose, the KF is parameterized using the predictions of $p_\text{radio}$, $v$, and (optionally) $acc$. The KF's measurement noise and process noise covariances are fine-tuned based on the predictions of $p_\text{radio}$, $v$, and (optional) $acc$. Again, the errors of KF are given in Section~\ref{sec:results} in Table~\ref{table:pose:results:data:results_accuracy_trajectory_sudden}.

\section{Results}
\label{sec:results}


\subsection{Effects of (Un)Directed Velocity}

\begin{table}[!t]
	\caption{Statistics of the dataset for evaluating the pose estimates.}
	\resizebox{\linewidth}{!}{%
	\setlength{\tabcolsep}{2pt}%
	\renewcommand{\arraystretch}{1.25}%
	\begin{tabular}{ l || r | r r | r r | r r | r r | r | r | r r r }
	Name & \parbox{0.75cm}{Subj.} & \multicolumn{2}{c|}{Total [$\#$]} & \multicolumn{2}{c|}{Training [$\#$]} & \multicolumn{2}{c|}{Valid. [$\#$]} & \multicolumn{2}{c|}{Test [$\#$]} & Duration & Distance & \multicolumn{3}{c}{$v_{ref}$ [\si{\meter\per\second}]}\\
	& [$\#$] & \rotatebox[origin=c]{90}{\parbox[c]{1.5cm}{\centering Segment}} & 
	\rotatebox[origin=c]{90}{\parbox[c]{1.5cm}{\centering Window/\\Feature}} & \rotatebox[origin=c]{90}{\parbox[c]{1.5cm}{\centering Segment}} & \rotatebox[origin=c]{90}{\parbox[c]{1.5cm}{\centering Window/\\Feature}} & \rotatebox[origin=c]{90}{\parbox[c]{1.5cm}{\centering Segment}} & \rotatebox[origin=c]{90}{\parbox[c]{1.5cm}{\centering Window/\\Feature}} & \rotatebox[origin=c]{90}{\parbox[c]{1.5cm}{\centering Segment}} & \rotatebox[origin=c]{90}{\parbox[c]{1.5cm}{\centering Window/\\Feature}} & [$min$]  & [\si{\kilo\meter}]  & $\varnothing$ & $min$ & $max$ \\\hline\hline
	\textbf{Accuracy} &  &  &  &  &  &  &  &  &  &  &  &  &  &   \\
	V1 & 20 & 160 & 112.500 & 112 & 78.750 & 16 & 11.250 & 32 & 22.500 & 1.203 & 185.40 &  2.5 & 0.8 & 7.9 \\
	V2 & 20 & 160 & 97.950 & 112 & 68.565 & 16 & 9.795 & 32 & 19.590 & 1.045 & 161.42 &  2.6 & 0.8 & 7.8 \\
	V3 & 20 & - & 31.200 & - & 21.840 & - & 3.120 & - & 6.240 & 666 & 71.9 &  1.8 &  0.8 & 3.6 \\\hline
	\textbf{Trajectory} &  &  &  &  &  &  &  &  &  &  &  &  &  &   \\
	Unknown Subj. & 2 & 16 & 9.795 & - & - & - & - & 16 & 9.795  & 105 & 16.14 &  2.5 &  0.7 & 3.7  \\
	\end{tabular}}
\label{table:pose:results:sequential:datasets}
\end{table}

\textbf{General Findings.} The general findings from the trajectory reconstruction are as follows: (1) No direct correlation was found between high velocity errors and high position errors for any of the methods. For example, all methods exhibited high velocity errors during the \textit{random} activity, yet the position errors were low for PDR, ML-GP, and RoNIN, or nearly linear for C/RNN and PDRNN (for a fair comparison, we trained PDRNN only on velocities). (2) Across all activities, the PDRNN method produced the smallest positional errors, followed by C/RNN. PDR yielded the worst results. (3) Smaller position errors correspond to better agreement between the MAE, $\text{CEP}_{95}$, MSE, and RMSE values, reflecting the distance between estimated positions and the reference, even without recalibration. In cases of significant outliers (high MSE or RMSE values), the reconstructed trajectories for PDR, ML-GP, and RoNIN exhibited considerable distortion, with noticeable deviations in shape from the reference trajectory. Instead, the trajectories for C/RNN and PDRNN showed better alignment, with smaller deviations from the reference. (4) As the subject's velocity increases, the reconstructed trajectories drift further from the reference. Notably, all methods performed well on the more complex pattern of the \textit{random} activity, with PDR, ML-GP, and RoNIN achieving lower position errors than during \textit{jogging} and \textit{running}. C/RNN and PDRNN showed superior performance. (5) Recalibration significantly reduced position errors (nearly halving them across all tests).

\textbf{Recalibration.} In a preliminary study we found that recalibrating the current position estimate $\hat{p}$ with the reference position improves the positional accuracy for all estimators across all error metrics. Shorter calibration intervals led to more accurate position estimates, with $1\,s$ intervals being sufficient to align $\hat{p}$ with $p_\text{ref}$. Calibration intervals exceeding $30\,s$ had no significant effect on the $f_s$ methods, while even $100\,s$ recalibration intervals enhanced the reconstruction errors for the C/RNN and PDRNN (trained on velocities only) methods. We select $30\,s$ intervals for the evaluation, as they mimic typical applications using GPS at a low update rate and reduce computational and communication load.

\begin{table}[!t]
	\caption{Position error of all methods of the trajectories in [$m$] of left out test subjects. Highest position accuracies are \textbf{bold}. "pure": a single initial position; "recalibrated": a single initial and a radio position every 30 $s$.}
	\label{table:pose:results:sequential:results_accuracy_trajectory}
	\setlength{\tabcolsep}{2pt}%
	\renewcommand{\arraystretch}{1.25}%
	\resizebox{\linewidth}{!}{%
		\begin{tabular}{ l l || r r r r | r r r r| r r r r| r r r r | r r r r } 
			\multicolumn{2}{l||}{Dataset} & \multicolumn{4}{c|}{PDR $K_g$} & \multicolumn{4}{c|}{ML-GP} & \multicolumn{4}{c|}{RoNIN} & \multicolumn{4}{c|}{C/RNN} & \multicolumn{4}{c}{PDRNN}\\
			\multicolumn{2}{l||}{\vtop{\hbox{\strut Left out}\hbox{\strut Subj.}}} & \rotatebox[origin=c]{90}{MAE} & \rotatebox[origin=c]{90}{MSE} & \rotatebox[origin=c]{90}{RMSE}  & \rotatebox[origin=c]{90}{$\text{CEP}_{95}$} 
			& \rotatebox[origin=c]{90}{MAE} & \rotatebox[origin=c]{90}{MSE} & \rotatebox[origin=c]{90}{RMSE} & \rotatebox[origin=c]{90}{$\text{CEP}_{95}$}  
			& \rotatebox[origin=c]{90}{MAE} & \rotatebox[origin=c]{90}{MSE} & \rotatebox[origin=c]{90}{RMSE}  & \rotatebox[origin=c]{90}{$\text{CEP}_{95}$}  
			& \rotatebox[origin=c]{90}{MAE} & \rotatebox[origin=c]{90}{MSE} & \rotatebox[origin=c]{90}{RMSE}  & \rotatebox[origin=c]{90}{$\text{CEP}_{95}$}  
			& \rotatebox[origin=c]{90}{MAE} & \rotatebox[origin=c]{90}{MSE} & \rotatebox[origin=c]{90}{RMSE}  & \rotatebox[origin=c]{90}{$\text{CEP}_{95}$}\\\hline\hline
			\multirow{4}{*}{\rotatebox[origin=c]{90}{\footnotesize pure}} 
			& Walking  & 4.28 & 25.51 & 4.84 & 8.01 & 3.15 & 22.88 & 3.93 & 7.18 & 2.41 & 10.05 & 2.77 & 4.45 & 0.71 & 0.87 & 0.93 & 1.83 & \textbf{0.59} & 0.49 & 0.70 & 1.19\\
			& Jogging  & 8.56 & 98.15 & 9.81 & 15.52 & 3.93 & 22.86 & 4.52 & 7.44 & 6.38 & 54.64 & 7.36 & 11.39 & 1.17 & 2.76 & 1.49 & 3.04 & \textbf{0.69} & 0.59 & 0.76 & 1.22\\
			& Running  & 9.09 & 120.54 & 9.84 & 14.42 & 5.70 & 42.05 & 6.44 & 9.94 & 2.74 & 12.65 & 3.55 & 6.61 & 1.20 & 2.18 & 1.34 & 2.06 & \textbf{0.95} & 1.36 & 1.13 & 2.07\\
			& Random & 4.50 & 34.32 & 5.38 & 9.31 & 2.88 & 13.56 & 3.56 & 6.71 & 1.87 & 5.77 & 2.35 & 4.67 & 1.52 & 4.11 & 1.86 & 3.98 & \textbf{0.92} & 1.37 & 1.12 & 2.16\\\hline
			\multirow{4}{*}{\rotatebox[origin=c]{90}{{\footnotesize recalibrated}}} 
			& Walking  & 2.38 & 9.42 & 3.04 & 5.88 & 1.67 & 6.61 & 2.38 & 4.84 & 1.47 & 3.22 & 1.79 & 3.54 & 0.57 & 0.54 & 0.74 & 1.60 & \textbf{0.36} & 0.23 & 0.47 & 0.93\\
			& Jogging  & 2.36 & 10.92 & 3.19 & 6.89 & 2.17 & 6.91 & 2.63 & 4.55 & 1.67 & 4.98 & 2.17 & 4.36 & 0.68 & 0.88 & 0.89 & 2.02 & \textbf{0.34} & 0.20 & 0.44 & 0.94\\
			& Running  & 3.13 & 18.18 & 4.14 & 8.43 & 2.23 & 8.97 & 2.97 & 6.79 & 1.73 & 5.08 & 2.24 & 4.39 & 0.75 & 0.85 & 0.92 & 1.89 & \textbf{0.33} & 0.22 & 0.46 & 1.00\\
			& Random  & 2.58 & 15.11 & 3.54 & 7.70 & 1.77 & 5.92 & 2.39 & 5.28 & 1.54 & 4.11 & 3.14 & 6.01 & 0.92 & 1.49 & 1.17 & 2.50 & \textbf{0.49} & 0.52 & 0.66 & 1.25
	\end{tabular}}
    \vspace{-0.5cm}
\end{table}

\begin{figure*}[!t]
    \centering
	\begin{minipage}[t]{0.195\linewidth}
        \centering
    	\includegraphics[trim=95 25 75 110, clip, width=1.0\linewidth]{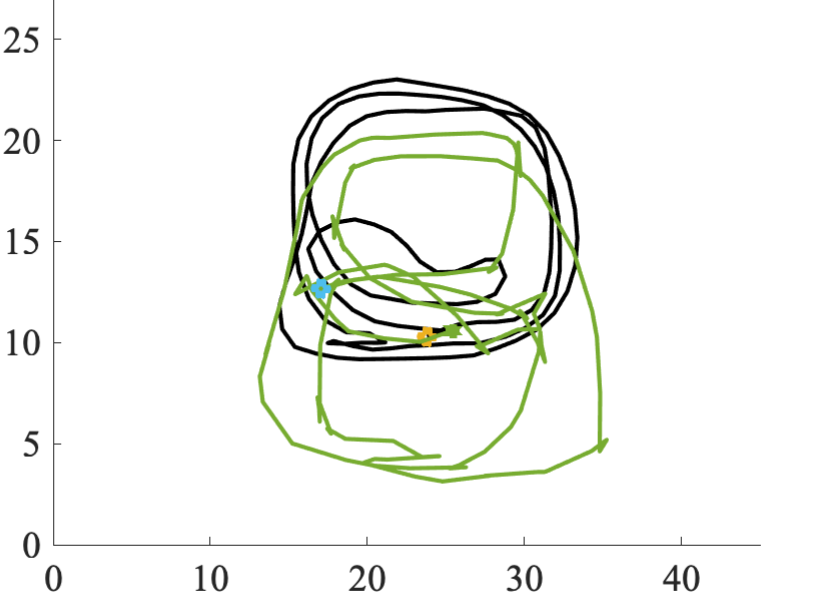}
        \subcaption{PDR.}
		\label{fig:pose:results:sequential:traj_walking_PDR_reset_0}
    \end{minipage}
    \hfill
	\begin{minipage}[t]{0.195\linewidth}
        \centering
    	\includegraphics[trim=95 25 75 110, clip, width=1.0\linewidth]{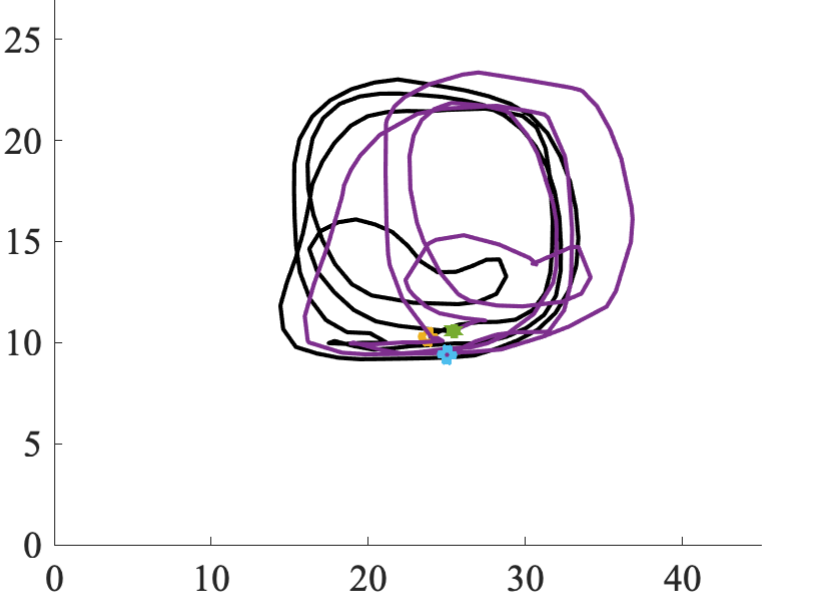}
        \subcaption{ML-GP.}
		\label{fig:pose:results:sequential:traj_walking_ML-GP_reset_0}
    \end{minipage}
    \hfill
	\begin{minipage}[t]{0.195\linewidth}
        \centering
    	\includegraphics[trim=95 25 75 110, clip, width=1.0\linewidth]{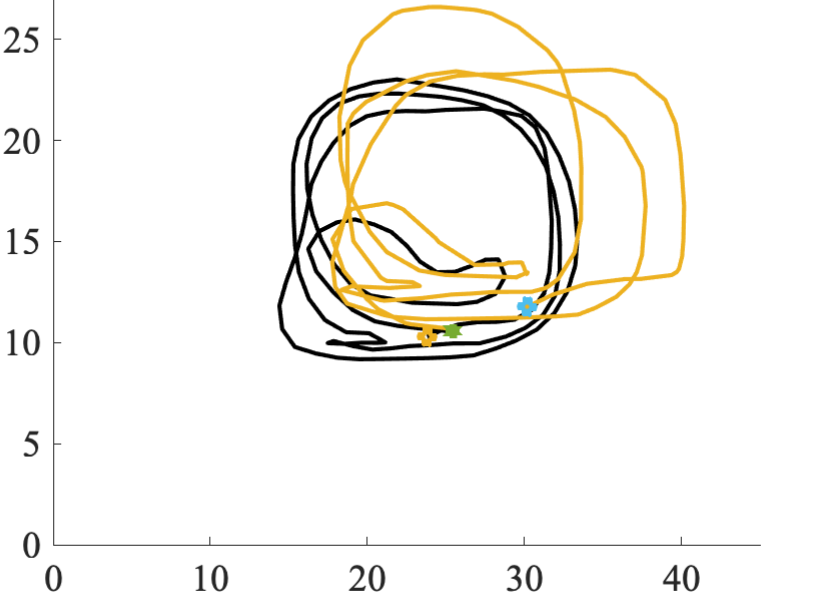}
        \subcaption{RoNIN.}
		\label{fig:pose:results:sequential:traj_walking_RoNIN_reset_0}
    \end{minipage}
    \hfill
	\begin{minipage}[t]{0.195\linewidth}
        \centering
    	\includegraphics[trim=95 25 75 110, clip, width=1.0\linewidth]{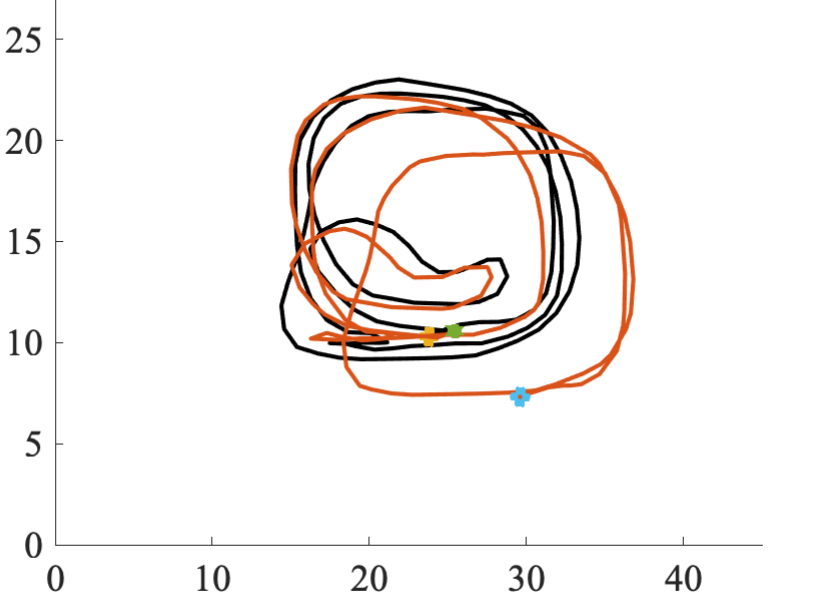}
        \subcaption{C/RNN.}
		\label{fig:pose:results:sequential:traj_walking_C-RNN_reset_0}
    \end{minipage}
    \hfill
	\begin{minipage}[t]{0.195\linewidth}
        \centering
    	\includegraphics[trim=95 25 75 110, clip, width=1.0\linewidth]{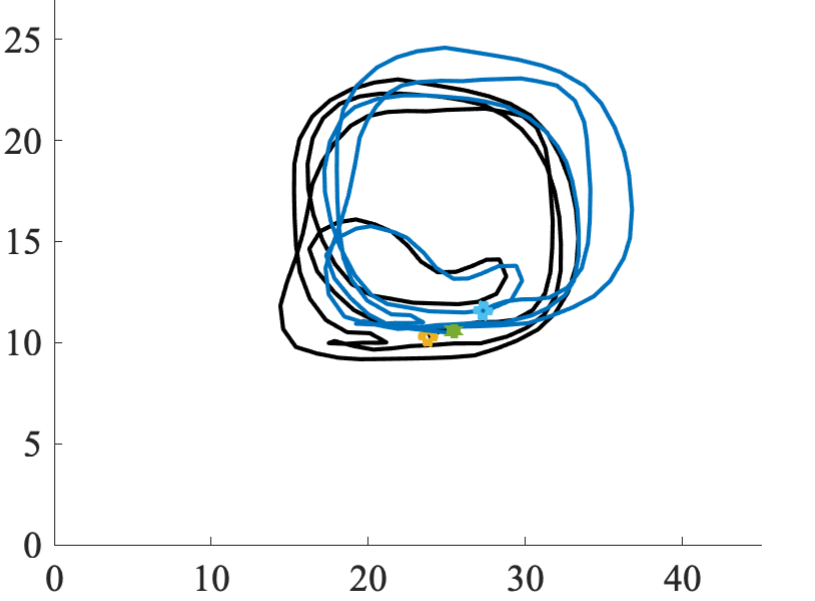}
        \subcaption{PDRNN.}
		\label{fig:pose:results:sequential:traj_walking_Hybrid_reset_0}
    \end{minipage}
	\begin{minipage}[t]{0.195\linewidth}
        \centering
    	\includegraphics[trim=190 30 25 115, clip, width=.9\linewidth]{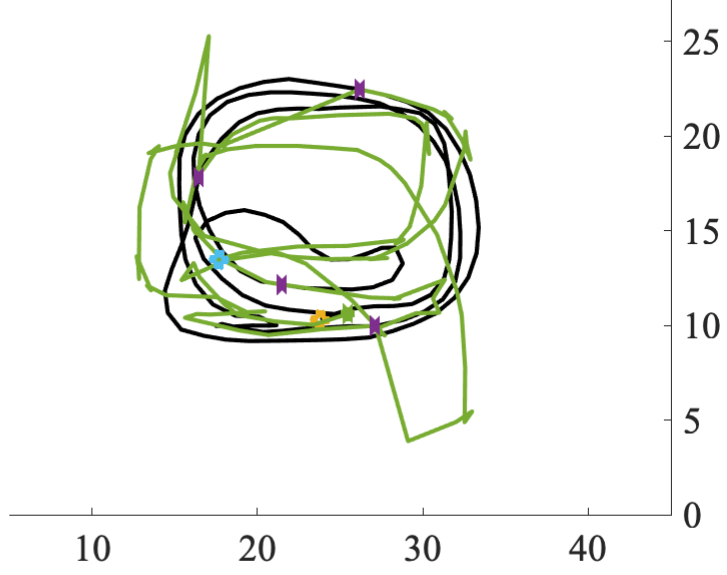}
        \subcaption{PDR, recalibration.}
        \label{fig:pose:results:sequential:traj_walking_PDR_reset_1}
    \end{minipage}
    \hfill
	\begin{minipage}[t]{0.195\linewidth}
        \centering
    	\includegraphics[trim=190 30 25 115, clip, width=.9\linewidth]{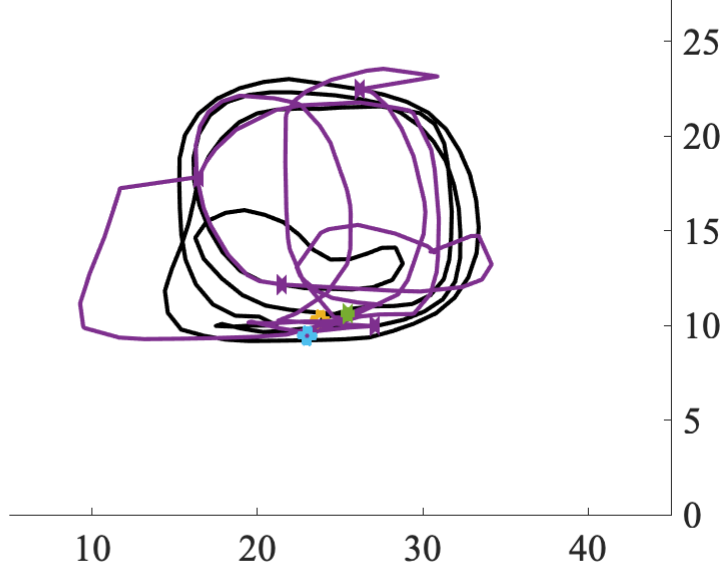}
        \subcaption{ML-GP, recalibration.}
		\label{fig:pose:results:sequential:traj_walking_ML-GP_reset_1}
    \end{minipage}
    \hfill
	\begin{minipage}[t]{0.195\linewidth}
        \centering
    	\includegraphics[trim=250 40 30 140, clip, width=.9\linewidth]{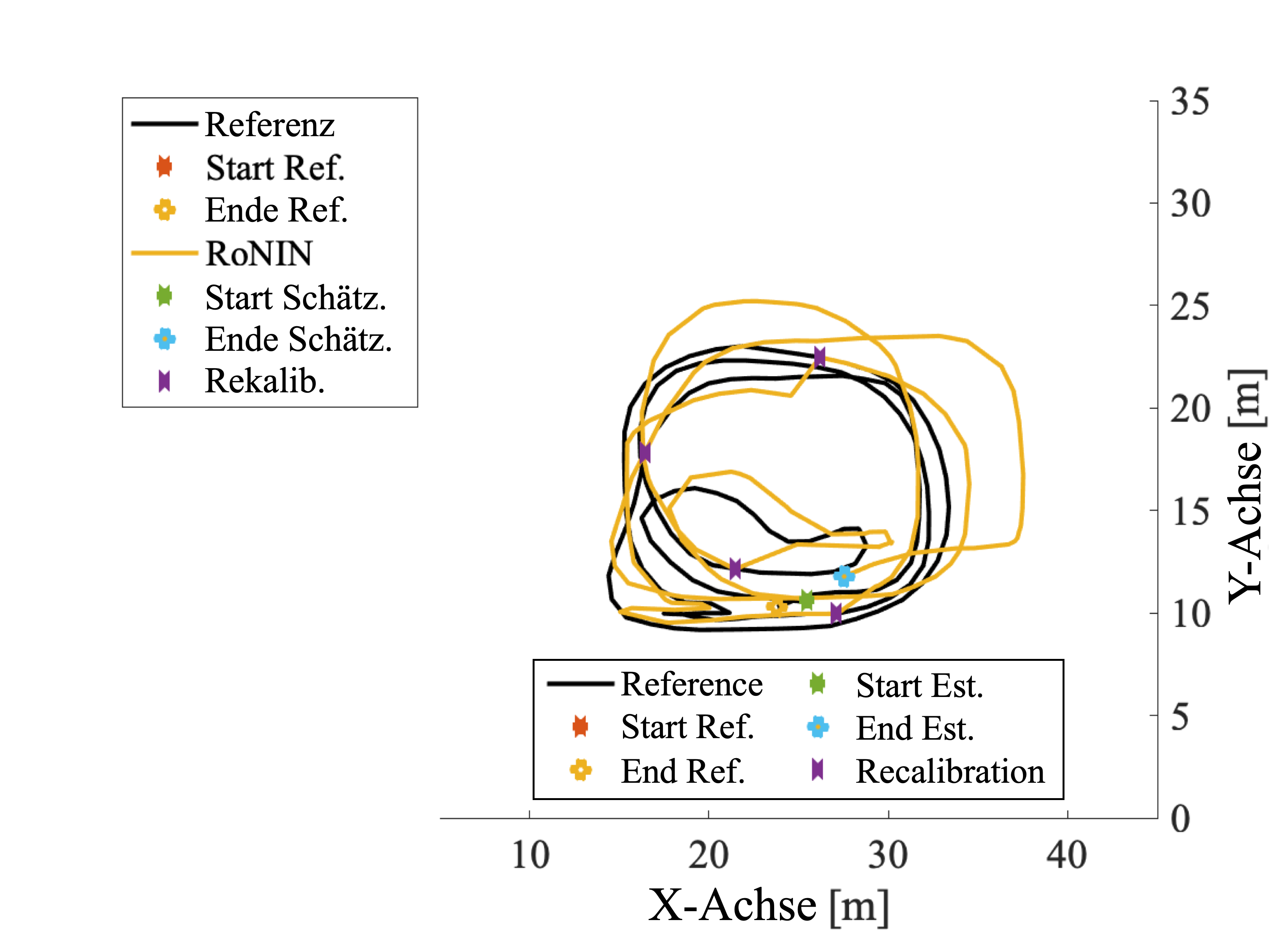}
        \subcaption{RoNIN, recalibration.}
		\label{fig:pose:results:sequential:traj_walking_RoNIN_reset_1}
    \end{minipage}
    \hfill
	\begin{minipage}[t]{0.195\linewidth}
        \centering
    	\includegraphics[trim=190 30 25 115, clip, width=.9\linewidth]{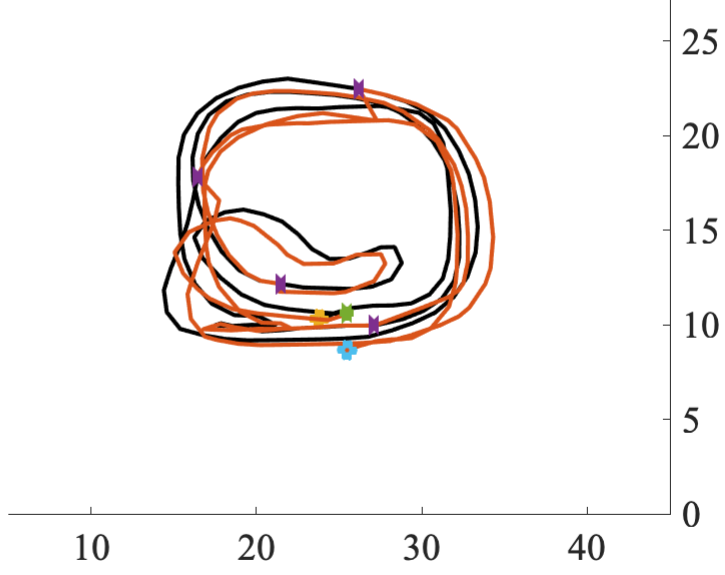}
        \subcaption{C/RNN, recalibration.}
		\label{fig:pose:results:sequential:traj_walking_C-RNN_reset_1}
    \end{minipage}
    \hfill
	\begin{minipage}[t]{0.195\linewidth}
        \centering
    	\includegraphics[trim=190 30 25 115, clip, width=.9\linewidth]{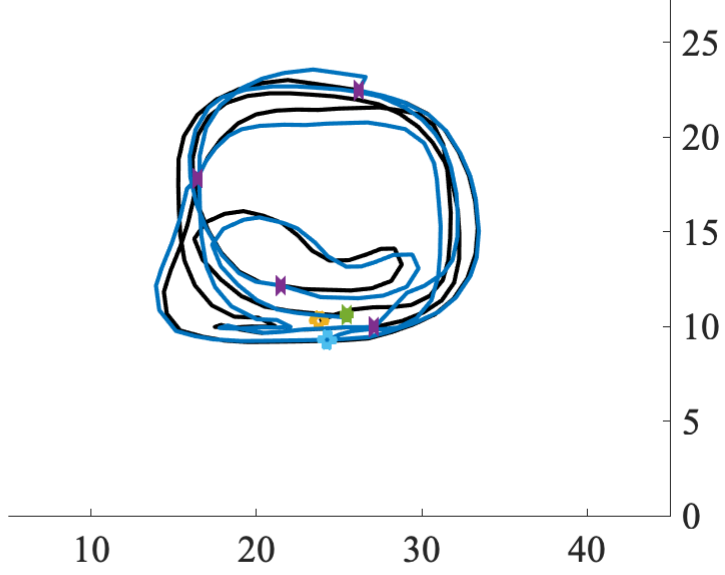}
        \subcaption{PDRNN, recalibration.}
		\label{fig:pose:results:sequential:traj_walking_Hybrid_reset_1}
    \end{minipage}
    \caption{Trajectory predictions of $3\,min$ from \textit{walking} of the left out test subject A (x- and y-axes in $m$ and colored lines represent the estimates. In line with Table 2, top row shows estimates of all methods using a single initial position "pure" and the bottom row shows methods when we recalibrate them on a radio position regularly. For a fair comparison we trained PDRNN without positions.}
	\vspace{-0.5cm}\label{fig:pose:results:sequential:results_traj_walking}
\end{figure*}

\textbf{Effects of Velocity on the Trajectory Reconstruction.} All methods produce the most accurate trajectory reconstruction during \textit{walking}. However, even for this low-velocity activity, PDR, ML-GP, and RoNIN exhibit deviations in the reconstructed trajectories compared to the reference, while the PDRNN method (trained on velocities only) reconstructs the trajectory with almost no error. As shown in Table~\ref{table:pose:results:sequential:results_accuracy_trajectory}, the errors and outliers progressively decrease from PDR to PDRNN. Larger outliers correlate with greater deviations from the reference trajectory. As velocity increases, the general ranking of the methods' reconstruction capabilities remains consistent, with PDR performing the worst and PDRNN yielding the best results. The drift is more noticeable with PDR, ML-GP, and RoNIN, compared to C/RNN and PDRNN, that demonstrate minimal drift. During \textit{jogging}, PDR, ML-GP, and RoNIN exhibit drift, typically undershooting the reference (estimated positions become smaller than the references). 

\textbf{Effects of Recalibration on the Trajectory Reconstruction Accuracy.} It is evident that recalibration significantly reduces position errors across all metrics for all activities. Generally, recalibration yields the greatest improvements when the velocity is low. As shown in Table~\ref{table:pose:results:sequential:results_accuracy_trajectory}(bottom row), recalibration nearly halves the position errors for PDR, ML-GP, and RoNIN, while its impact is less pronounced for C/RNN and PDRNN (trained only on velocities) methods. Furthermore, PDR, ML-GP, and RoNIN, that exhibit the most drift without recalibration, show the greatest improvements. Although recalibration notably enhances trajectory reconstruction for PDR, ML-GP, C/RNN, and PDRNN during the \textit{random} activity, RoNIN still experiences significant position errors. This may be due to RoNIN's inability to sufficiently learn the dependencies between different movements and velocities, leading to the misprocessing of large outliers. Given that C/RNN and PDRNN perform well with or without recalibration, it appears that ML-based methods are more effective at handling a wide range of velocities.

\textbf{Conclusion.} As velocity increases, drift in the reconstructed trajectory (only on velocity data) becomes more pronounced, accompanied by position jumps, wobbling, and jitter, particularly with frequent changes in direction. Consequently, the reconstruction of the trajectory performs best when velocity changes moderately, allowing for recovery from rapid movement and drift, while minimizing directional fluctuations. By incorporating recalibrations, the estimated trajectories are brought closer to the reference, even for C/RNN and PDRNN, that already perform well without recalibration. It appears that C/RNN and PDRNN excel in three key areas: (1) effectively handling varying velocities and the corresponding signal noise, as evidenced by consistently low error metrics across all activities; (2) accurately interpreting velocity scaling, that enables precise position estimates even during rapid motion changes; and (3) efficiently de-noising the inputs, as reflected in their lower errors.

\subsection{Evaluation of PDRNN}

\begin{table}[!t]
	\caption{Reconstruction error [$m$] of PDRNN on the unknown subjects with varying input data streams.}
	\label{table:pose:results:data:results_inputvari}
	\centering
	\scriptsize \begin{tabular}{ l || r r r r } & \multicolumn{4}{c}{PDRNN}\\
	\vtop{\hbox{\strut Input-}\hbox{\strut variation}} & \rotatebox[origin=c]{90}{MAE} & \rotatebox[origin=c]{90}{MSE} & \rotatebox[origin=c]{90}{RMSE} & \rotatebox[origin=c]{90}{$\text{CEP}_{95}$}\\\hline\hline
    $p_\text{radio}, v$ & \textbf{0.0375} & \textbf{0.0141} & \textbf{0.0027} & \textbf{0.0991}\\
    $p_\text{radio}, acc$ & 0.2722 & 0.0286 & 0.0421 & 0.4356\\ 
    $p_\text{radio}, \theta_\text{ori}$ & 0.2156 & 0.0198 & 0.0375 & 0.2911\\
    $p_\text{radio}, v, acc$ & 0.1144 & 0.0243 & 0.0324 & 0.2007\\
    $p_\text{radio}, v, \theta_\text{ori}$ & 0.1377 & 0.0212 & 0.0371 & 0.2289\\
    $p_\text{radio}, acc, \theta_\text{ori}$ & 0.3481 & 0.0463 & 0.1269 & 0.4108\\
    $p_\text{radio}, v, acc, \theta_\text{ori}$ & 0.1962 & 0.0342 & 0.0452 & 0.2577
	\end{tabular}
    \vspace{-0.5cm}
\end{table}

\begin{figure}[!t]
    \centering
    \includegraphics[trim={85 10 75 25},clip,width=1\linewidth]{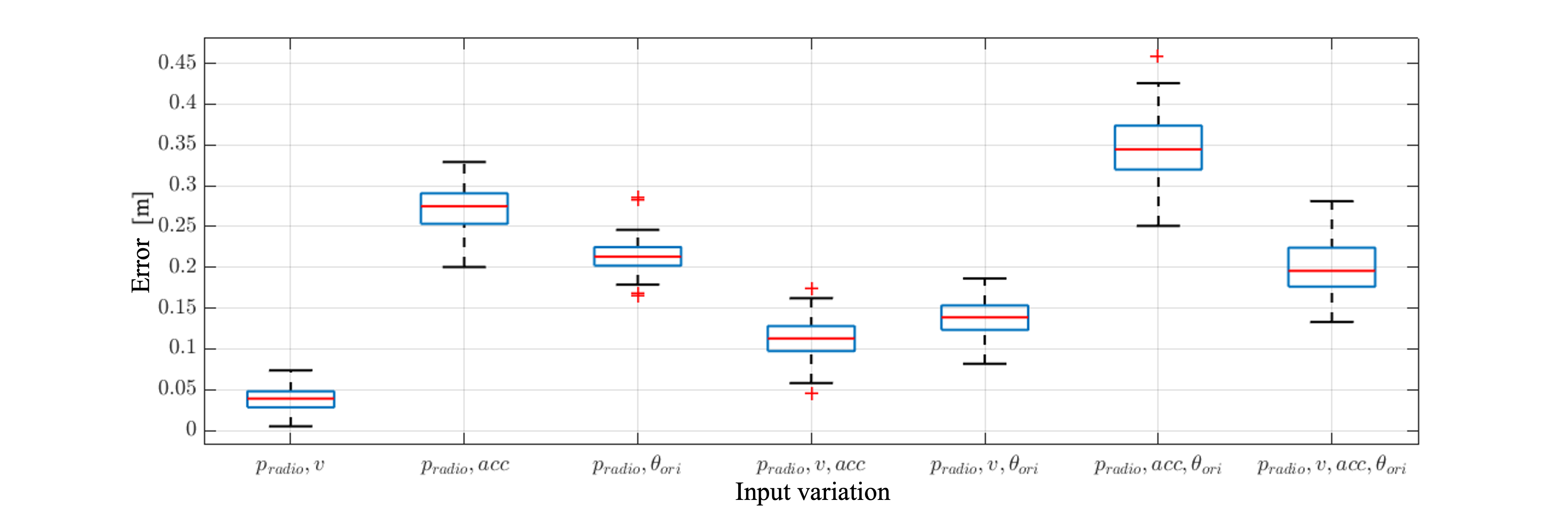}
    \caption{PDRNN pose estimation accuracy for various input data. (MAE = red line; outliers: red $ + $; and standard deviation).}
    \label{fig:pose:results:data:05-io-combinations}
    \vspace{-0.5cm}
\end{figure}

\textbf{Effect of Input Variations.} The input combination significantly impacts model accuracy. Models incorporating both radio position ($p_\text{radio}$) and velocity ($v$) outperform those without $v$ (e.g., $p_\text{radio}$, $acc$; $p_\text{radio}, \theta_\text{ori}$; $p_\text{radio}, acc, \theta_\text{ori}$). Including acceleration ($acc$) generally leads to poorer performance. Even when acceleration is noiseless (using a moving average filter, $\text{SMA}=10$), the poses remain inaccurate. The combination of $p_\text{radio}$ and $v$ achieves the highest positional accuracy ($\text{MAE} = 0.0375$, $\text{MSE} = 0.0141$, $\text{RMSE} = 0.0027$, $\text{CEP}_{95} = 0.0991$ in $m$). Instead, the combination of $p_\text{radio}$ with the signal magnitude vector of acceleration ($p_\text{radio}, acc$) yields the worst results ($\text{MAE} = 0.2722$, $\text{MSE} = 0.0286$, $\text{RMSE} = 0.0421$, $\text{CEP}_{95} = 0.4356$), and combining radio position, velocity, and acceleration ($p_\text{radio}, v, acc$) leads to worse results than $p_\text{radio}, v$ alone ($\text{MAE} = 0.1144$, $\text{MSE} = 0.0243$, $\text{RMSE} = 0.0324$, $\text{CEP}_{95} = 0.2007$). Adding acceleration contributes little beyond velocity and may even hinder accuracy, as no correlation exists between position and acceleration when velocity is absent. The KF with a constant velocity motion model provides significantly more accurate poses than with a constant acceleration model, even when $acc$ is added. The inaccuracy with $acc$ may stem from measurement noise, that the KF cannot reliably define for unknown data. The KF, optimized for $p_\text{radio}, v$, achieves $\text{MAE} = 0.3244$, $\text{MSE} = 0.0411$, $\text{RMSE} = 0.0573$, and $\text{CEP}_{95} = 0.6019$ meters, though less accurate than PDRNN but more accurate than the model-based PDR method ($\text{MAE} = 0.4900$, $\text{MSE} = 0.5200$, $\text{RMSE} = 0.6600$, $\text{CEP}_{95} = 1.2500$ meters). The KF performs best when optimized for the dataset but suffers higher errors with data from an unknown test subject due to dataset differences. Adding orientation ($\theta_\text{ori}$) to position generally worsens results, with the combination of $p_\text{radio}, v, \theta_\text{ori}$ yielding $\text{MAE} = 0.1377$, $\text{MSE} = 0.0212$, $\text{RMSE} = 0.0371$, and $\text{CEP}_{95} = 0.2289$ meters. This may be due to the radio and inertial sensors being in the same coordinate system, where the radio position implicitly indicates orientation, leading to estimation errors that confuse PDRNN. Therefore, orientation should only be integrated into the model if its error variance is low and it does not degrade position accuracy. However, when radio and inertial sensors are in different coordinate systems, orientation should be included.

\begin{table}[!t]
	\caption{Reconstruction errors of the PDRNN method of the trajectories [\si{\meter}] of the unknown test subjects with a varying forecast horizon.}
	\label{table:pose:results:data:results_horizon}
	\centering
	\scriptsize \begin{tabular}{ r r || r r r r } 
	   \multicolumn{2}{c}{Configuration} & \multicolumn{4}{c}{PDRNN}\\
		\vtop{\hbox{\strut Forecast-}\hbox{\strut Horizon [$s$]}} & \vtop{\hbox{\strut Sequence-}\hbox{\strut length [$s$]}} & \rotatebox[origin=c]{90}{MAE} & \rotatebox[origin=c]{90}{MSE} & \rotatebox[origin=c]{90}{RMSE} & \rotatebox[origin=c]{90}{$\text{CEP}_{95}$}\\\hline\hline
		0.0 & 0.64 & 0.0651 & 0.0227 & 0.0229 & 0.0919 \\
        0.0 & 1.28 & \textbf{0.0375} & 0.0141 & 0.0027 & 0.0991 \\
        0.0 & 2.56 & 0.1066 & 0.0341 & 0.0975 & 0.2356 \\\hline
        1.0 & 0.64 & 0.0667 & 0.0243 & 0.0324 & 0.1319 \\
        1.0 & 1.28 & \textbf{0.0489} & 0.0076 & 0.0057 & 0.1074 \\
        1.0 & 2.56 & 0.1175 & 0.0546 & 0.0975 & 0.2356 \\\hline
        2.0 & 0.64 & 0.4916 & 0.1966 & 0.2841 & 0.9835 \\
        2.0 & 1.28 & \textbf{0.3775} & 0.1745 & 0.2793 & 0.4745 \\
        2.0 & 2.56 & 0.5733 & 0.2981 & 0.4129 & 1.3213 \\\hline
        3.0 & 0.64 & 8.4357 & 3.9453 & - & - \\
        3.0 & 1.28 & \textbf{3.8741} & 1.4765 & - & - \\
        3.0 & 2.56 & 10.3233 & 4.9989 & - & -
	\end{tabular}
    \vspace{-0.5cm}
\end{table}

\begin{figure}[!t]
    \centering
    \includegraphics[trim=2.25cm 2.5cm 2.5cm 2.5cm, clip,width=\linewidth]{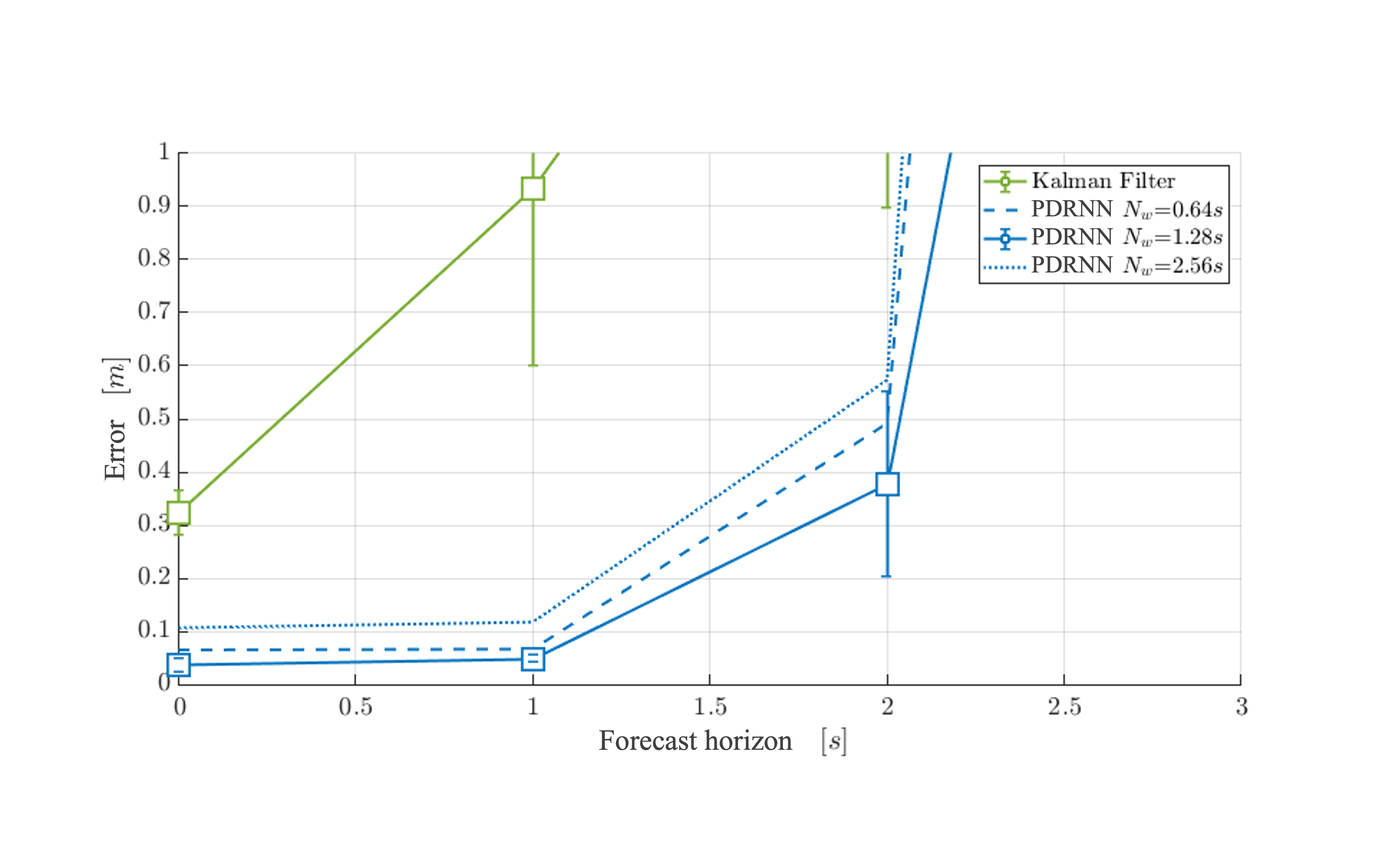}
    \caption{Pose accuracy of PDRNN (in $m$) trained on $p_\text{radio}$, $v$ over the forecast horizon (from $0\,s$ to $3\,s$) for different sequence lengths ($N_w \in [0.64, 1.28, 2.56]\,s$). The lines visualize the error (MAE) and the length of the vertical lines of the boxes represents the degree of error variance.}
    \label{fig:pose:results:data:05-lookahead}
    \vspace{-0.5cm}
\end{figure}

\textbf{Effect of the Forecast Horizon.} Figure~\ref{fig:pose:results:data:05-lookahead} shows pose accuracy in relation to the forecast horizon, with error variance presented for the best PDRNN model (trained on $p_\text{radio}$, $v$) and KF. The KF generally exhibits higher inaccuracies, tripling pose errors with a $1\,s$ prediction horizon. As the KF uses only the last timestep to predict the next, the sequence length does not influence its accuracy. This limited knowledge may explain why KF performs poorly for future pose predictions. PDRNN, however, yields the most accurate predictions with a sequence length of 128 timesteps and a $1\,s$ forecast horizon, compensating for measurement and system delays ($1\,s$ forecast with sequence length $1.28\,s$: $\text{MAE} = 0.0489$, $\text{MSE} = 0.0076$, $\text{RMSE} = 0.0057$, $\text{CEP}_{95} = 0.1074$ meters). Shorter sequences ($0.64\,s$) are less accurate ($1\,s$ forecast with $0.64\,s$ sequence: $\text{MAE} = 0.0667$, $\text{MSE} = 0.0243$, $\text{RMSE} = 0.0324$, $\text{CEP}_{95} = 0.1319$ meters). Instead, long sequences (e.g., $2.56\,s$) negatively affect accuracy ($1\,s$ forecast with $2.56\,s$ sequence: $\text{MAE} = 0.1175$, $\text{MSE} = 0.0546$, $\text{RMSE} = 0.0975$, $\text{CEP}_{95} = 0.2356$ meters). Excess information from long sequences, such as circular and spiral motions, may cause PDRNN to focus on unnecessary details, leading to decreased accuracy. We also observed that forecast horizons over $2\,s$ yield significantly worse accuracy ($2\,s$ forecast with $1.28\,s$ sequence: $\text{MAE} = 0.3775$, $\text{MSE} = 0.1745$, $\text{RMSE} = 0.2793$, $\text{CEP}_{95} = 0.4745$ meters), and horizons over $3\,s$ result in implausible pose accuracy, leading to the exclusion of tests with forecasts $\geq 2\,s$. For forecast horizons $< 1\,s$, error increases almost linearly with the horizon length. In an independent experiment, long input sequences ($30\,s$) of simple circular trajectories allowed accurate $10\,s$ future pose predictions. Instead, short input sequences ($1.28\,s$) produced larger errors for $10\,s$ predictions. This may be because long sequences fully capture circular motions, while short sequences can only learn partial circles. The lack of world knowledge about circular motions in shorter sequences prevents accurate long-term predictions. Notably, the LSTM's context vector seems to provide enough capacity for learning simple circular motions over long trajectories, indicating a strong correlation between accuracy, sequence length, LSTM capacity, and forecast horizon.

\begin{figure}[!t]
    \centering
	\begin{minipage}[t]{0.493\linewidth}
        \centering
    	\includegraphics[trim=2.5cm 0 2.5cm 0, clip, width=1.0\linewidth]{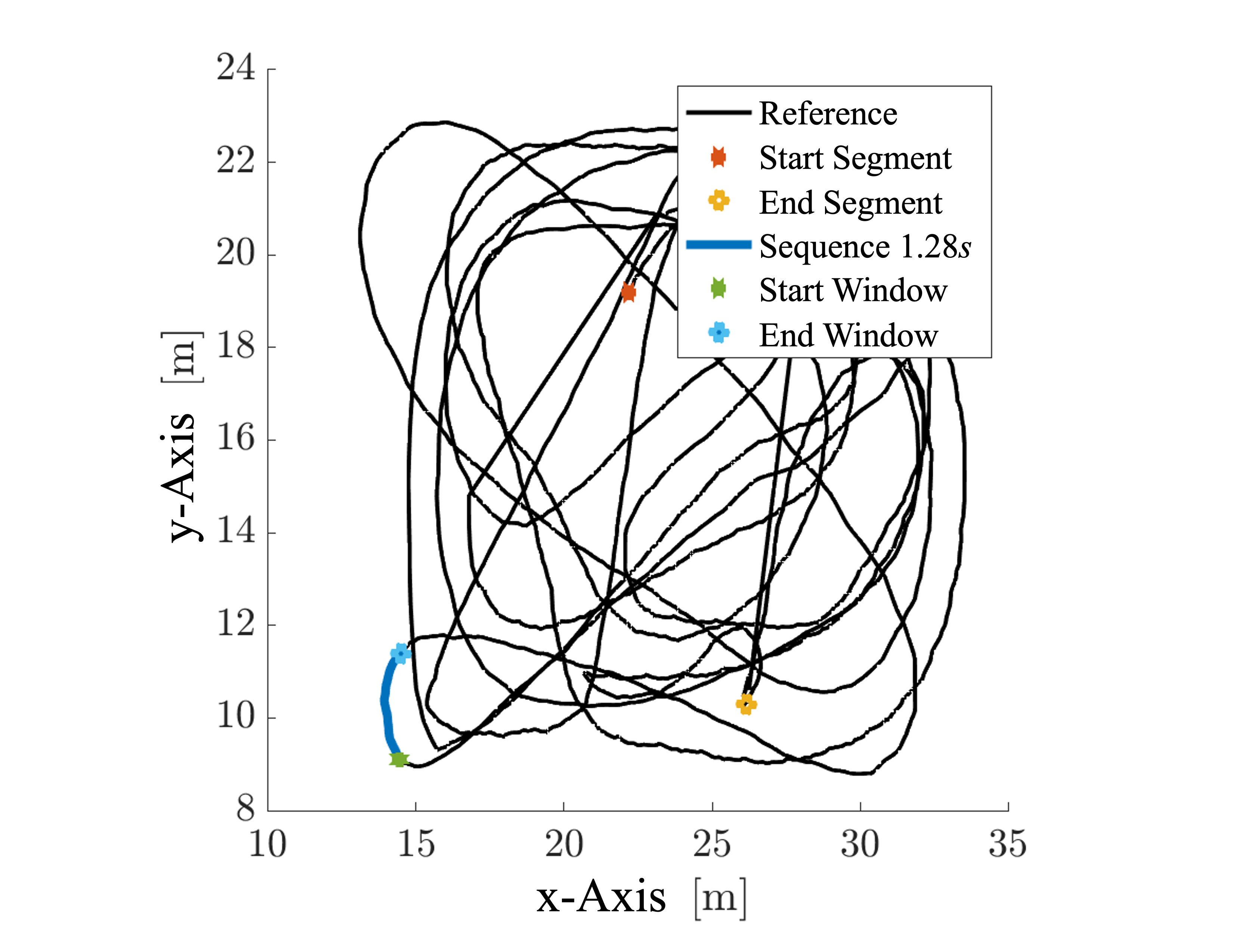}
        \subcaption{Short sequence 1.28~\si{\s}.}
    \end{minipage}
    \hfill
	\begin{minipage}[t]{0.493\linewidth}
        \centering
    	\includegraphics[trim=2.5cm 0 2.5cm 0, clip, width=1.0\linewidth]{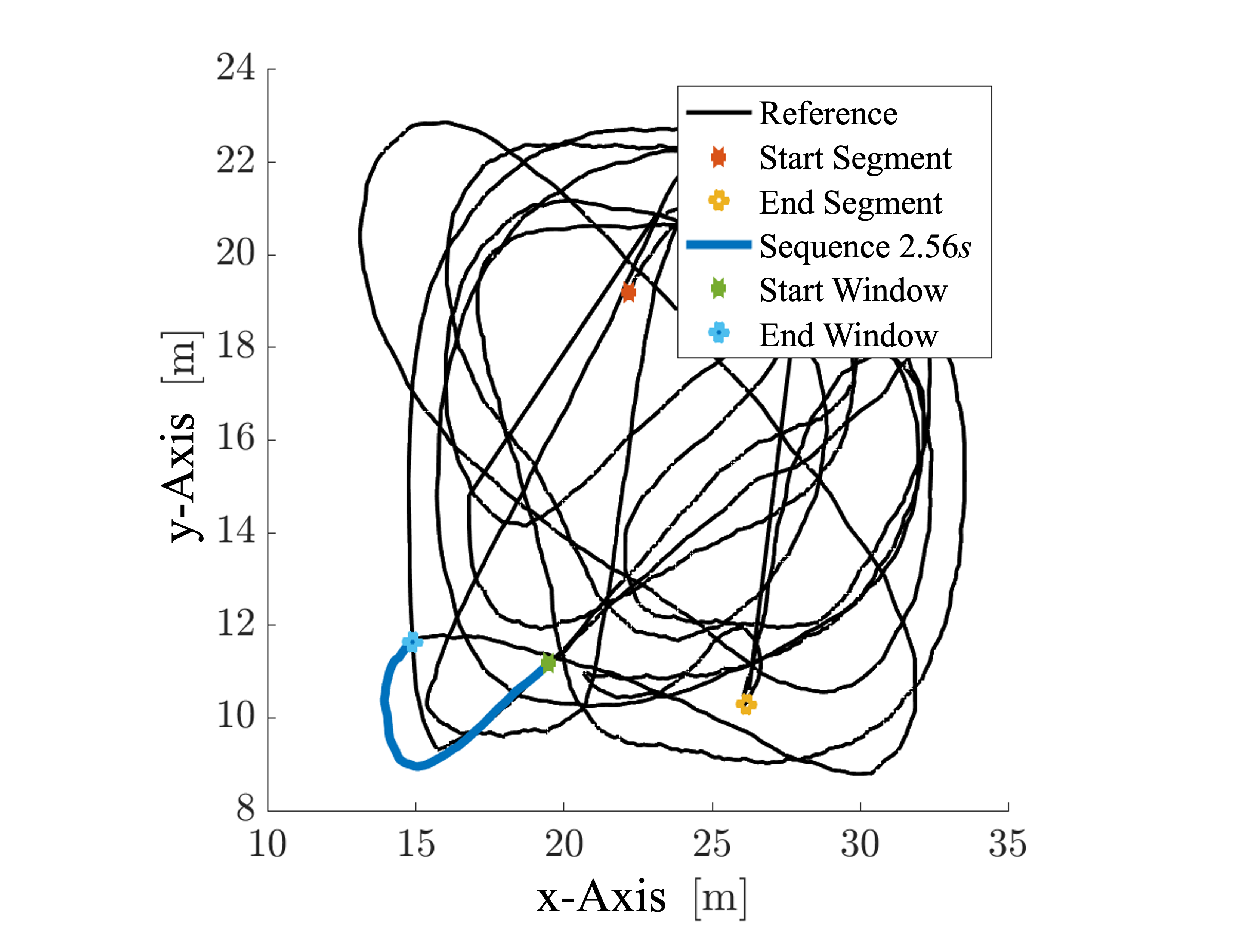}
        \subcaption{Long sequence 2.56~\si{\s}.}
    \end{minipage}
    \caption{Effects of the sequence length (blue line) on generalizability. Shorter sequences (left) are more likely to be contained in more complex trajectories and are more likely to reconstruct them. Longer sequences (right) are often unique and therefore more difficult to generalize, i.e., the blue sequence (right) is only found once in the entire segment, so that the motion model cannot use this movement for the reconstruction.}
    \label{fig:pose:results:data:sequence}
    \vspace{-0.125cm}
\end{figure}

\textbf{Effect of Sequence Length.} The pose accuracy of PDRNN (trained on $p_\text{radio}$, $v$) improves with sequence lengths up to 128 values, suggesting that the context vector approaches its maximum information capacity at this point (sequence length = 128: $\text{MAE} = 0.0375$, $\text{MSE} = 0.0141$, $\text{RMSE} = 0.0027$, $\text{CEP}_{95} = 0.0991$ meters). Beyond this length, accuracy declines, likely due to the context vector being overloaded (sequence length = 256: $\text{MAE} = 0.1012$, $\text{MSE} = 0.0537$, $\text{RMSE} = 0.0646$, $\text{CEP}_{95} = 0.1944$ meters). Shorter sequences with 64 values also yield poorer positional accuracy compared to 128 values (sequence length = 64: $\text{MAE} = 0.0615$, $\text{MSE} = 0.0034$, $\text{RMSE} = 0.0431$, $\text{CEP}_{95} = 0.1254$ meters). This may be attributed to insufficient information in the smaller window for the velocity estimate, leading to inaccuracies that negatively impact pose estimation. Short sequences often capture simple short, curvy, and straight movements, whereas longer sequences include more complex motion patterns. Longer trajectories may be reconstructed from combinations of short curves and straight lines, enabling models trained on shorter sequences to predict longer, complex motions accurately. Conversely, trajectory shapes in longer sequences tend to be more specialized and restrictive, making it harder to reconstruct unfamiliar shapes (sequence length = 512: $\text{MAE} = 0.2641$, $\text{MSE} = 0.1123$, $\text{RMSE} = 0.1435$, $\text{CEP}_{95} = 0.3546$ meters). Figure~\ref{fig:pose:results:data:sequence} illustrates examples of simple (left) and complex (right) motion patterns.

\begin{figure}[!t]
    \centering
	\begin{minipage}[t]{0.493\linewidth}
        \centering
    	\includegraphics[trim=2.5cm 0 2.5cm 0, clip, width=1.0\linewidth]{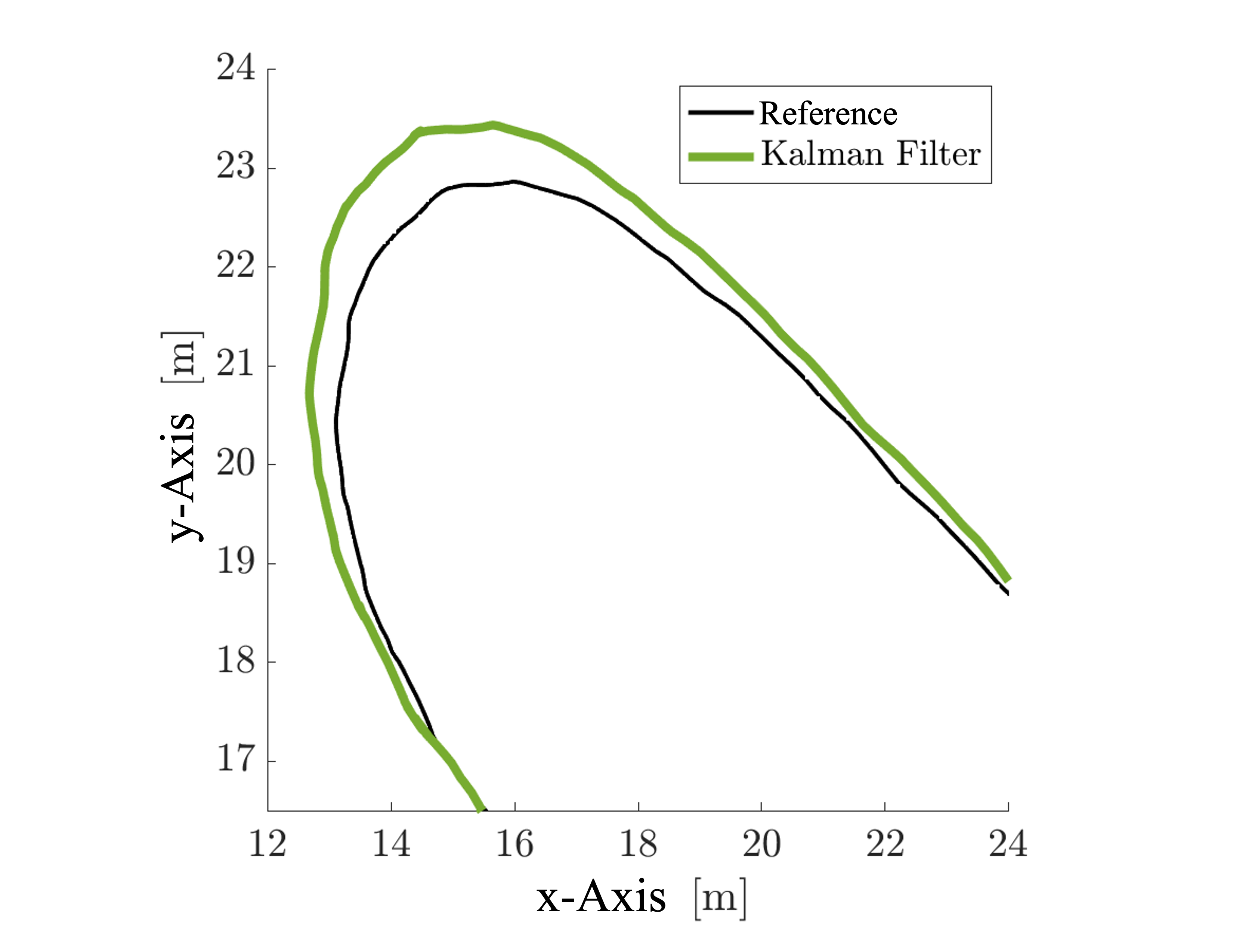}
        \subcaption{KF.}
    \end{minipage}
    \hfill
	\begin{minipage}[t]{0.493\linewidth}
        \centering
    	\includegraphics[trim=2.5cm 0 2.5cm 0, clip, width=1.0\linewidth]{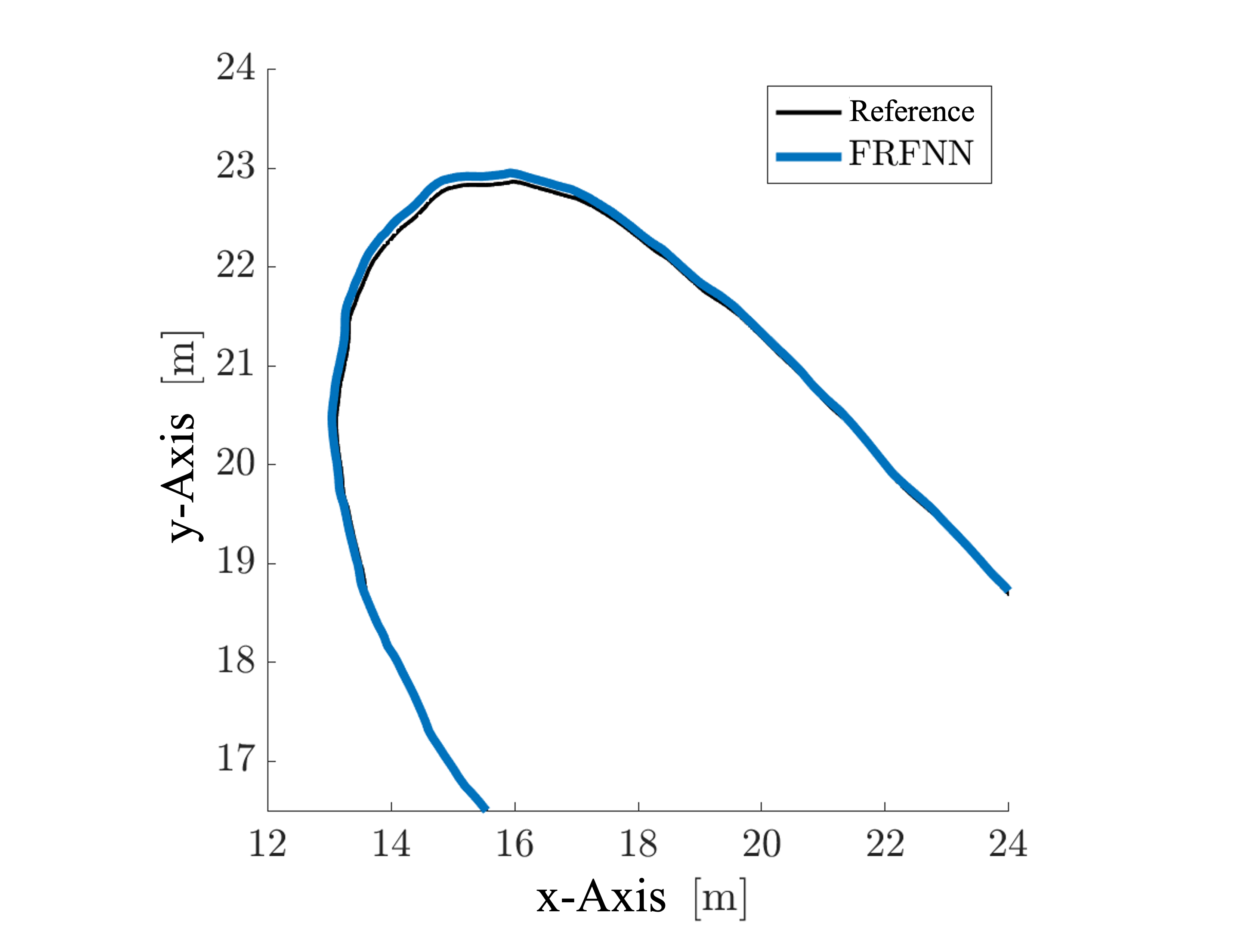}
        \subcaption{PDRNN.}
    \end{minipage}
    \caption{Positional accuracy of an example trajectory for the previously unseen subject A (random movement) with abrupt directional changes is depicted. The predictions of the KF and PDRNN (trained on $p_\text{radio}$, $v$) are compared during direction changes. Initially ($x = 15\,m$, $y = 17\,m$), the KF accurately models the position. However, following a change in direction, significant errors appear in the KF predictions ($x = 17\,m$, $y = 22.5\,m$), with recovery only after $3.8\,s$ ($x = 24\,m$, $y = 19\,m$). Instead, PDRNN maintains error-free predictions during directional changes.} \label{fig:pose:results:data:suddensequence}
    \vspace{-0.5cm}
\end{figure}

\begin{table}[!t]
	\caption{Reconstruction error of the KF and PDRNN method of the trajectories in [\si{\meter}] of the unknown subjects.}
	\resizebox{\linewidth}{!}{%
		\begin{tabular}{ l l || r r r r | r r r r } 
			\multicolumn{2}{l||}{Dataset} & \multicolumn{4}{c|}{KF} & \multicolumn{4}{c}{PDRNN}\\
			\multicolumn{2}{l||}{Left out Subj.} 
			& \rotatebox[origin=c]{90}{MAE} & \rotatebox[origin=c]{90}{MSE} & \rotatebox[origin=c]{90}{RMSE}  & \rotatebox[origin=c]{90}{$\text{CEP}_{95}$} 
			& \rotatebox[origin=c]{90}{MAE} & \rotatebox[origin=c]{90}{MSE} & \rotatebox[origin=c]{90}{RMSE} & \rotatebox[origin=c]{90}{$\text{CEP}_{95}$}\\\hline\hline
			\multirow{4}{*}{\rotatebox[origin=c]{90}{}} 
			& Walking    & 0.1832 & 0.0536 & 0.2225 & 0.3813 & \textbf{0.0312} & 0.0013 & 0.0355 & 0.0588 \\
			& Jogging   & 0.1917 & 0.0603 & 0.2339 & 0.4264 & \textbf{0.0378} & 0.0022 & 0.0454 & 0.0780 \\
			& Running   & 0.1970 & 0.0569 & 0.2377 & 0.4378 & \textbf{0.0536} & 0.0042 & 0.0594 & 0.0925 \\
			& Random & 0.2514 & 0.0895 & 0.2853 & 0.4595 & \textbf{0.0789} & 0.00072 & 0.0892 & 0.1367
	\end{tabular}}
	\label{table:pose:results:data:results_accuracy_trajectory_sudden}
    \vspace{-0.5cm}
\end{table}

\begin{figure}[!t]
    \centering
	\begin{minipage}[t]{0.493\linewidth}
        \centering
    	\includegraphics[trim=45 45 50 0, clip, height=3.25cm]{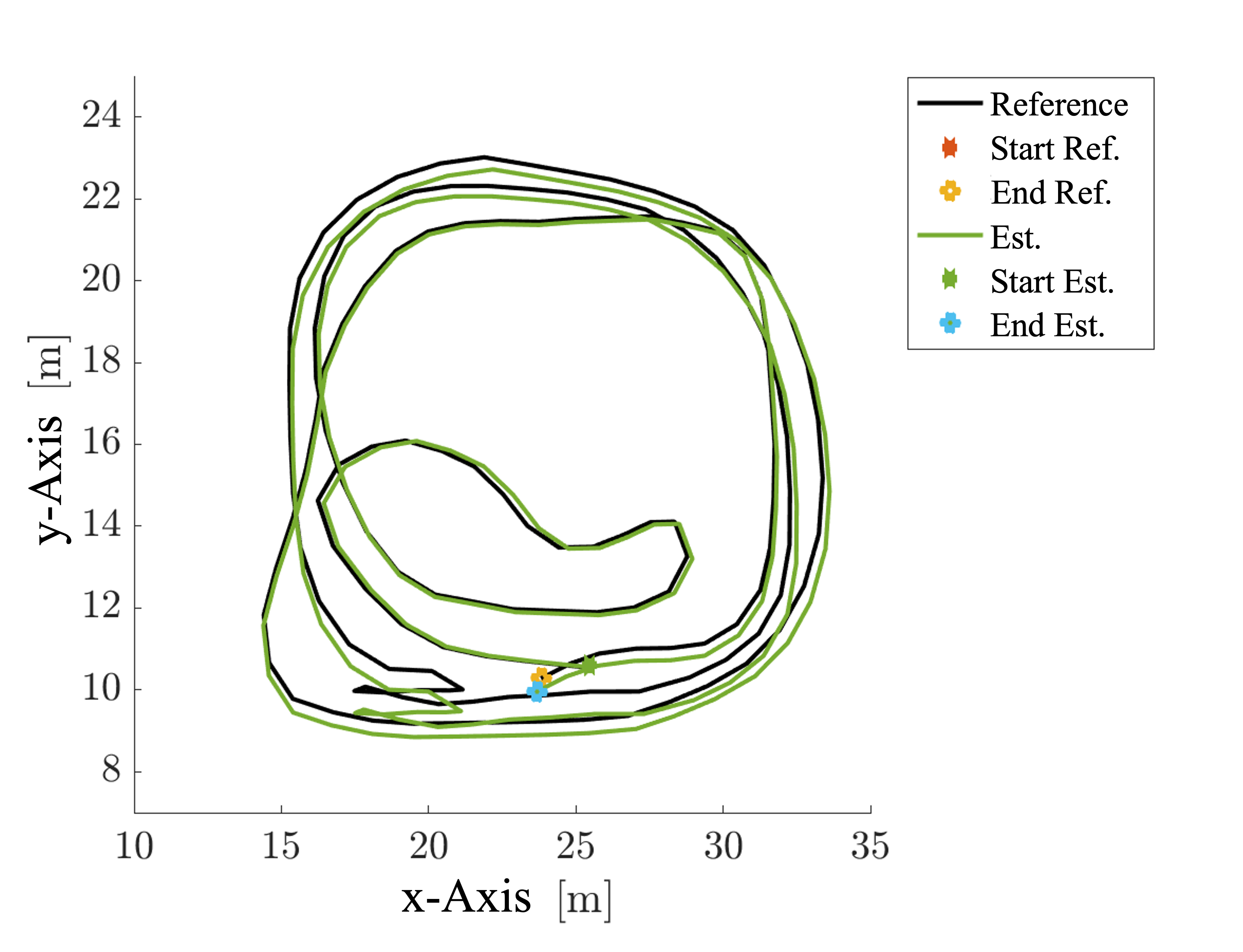}
        \subcaption{KF, \textit{walking}.}
		\label{fig:pose:results:data:walking_KF}
    \end{minipage}
    \hfill
	\begin{minipage}[t]{0.493\linewidth}
        \centering
    	\includegraphics[trim=30 35 180 0, clip, height=3.25cm]{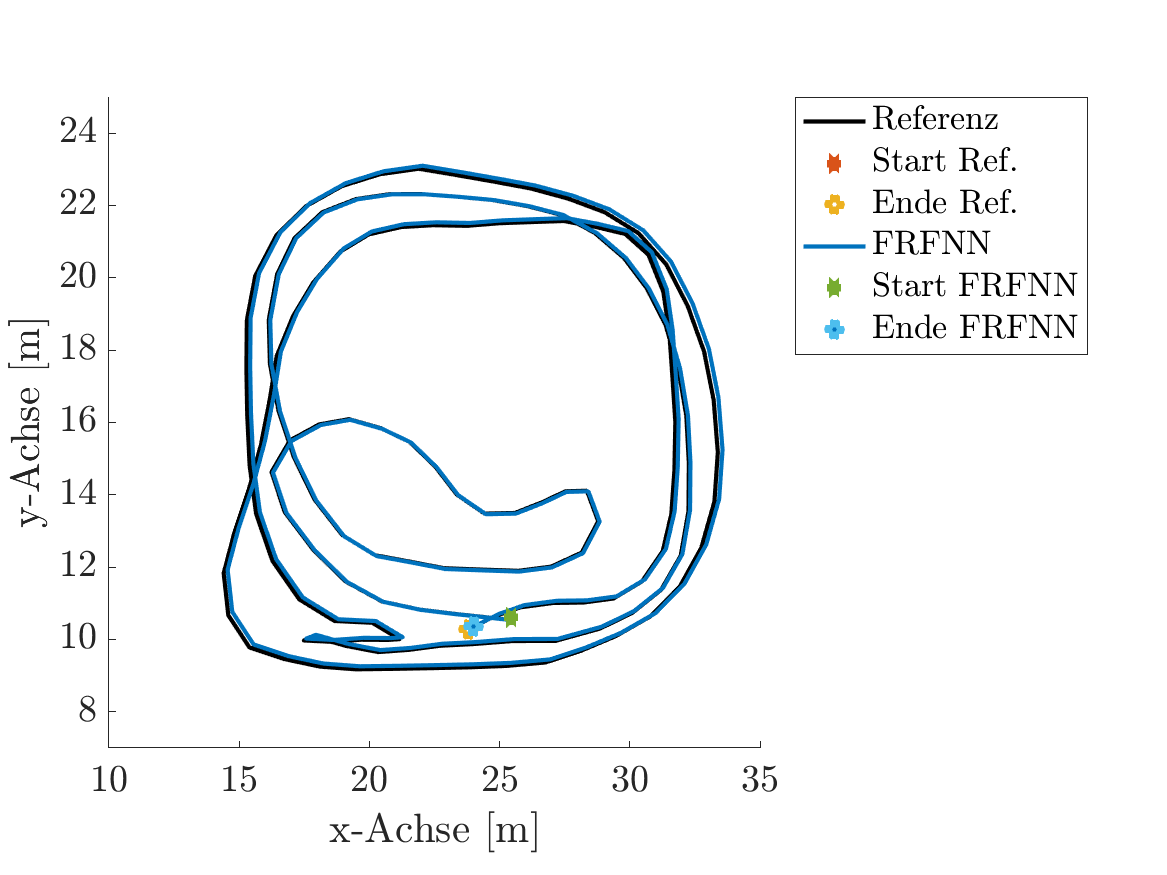}
        \subcaption{PDRNN, \textit{walking}.}
		\label{fig:pose:results:data:walking_PDRNN}
    \end{minipage}
    \hfill
	\begin{minipage}[t]{0.493\linewidth}
        \centering
    	\includegraphics[trim=30 25 160 0, clip, height=3.25cm]{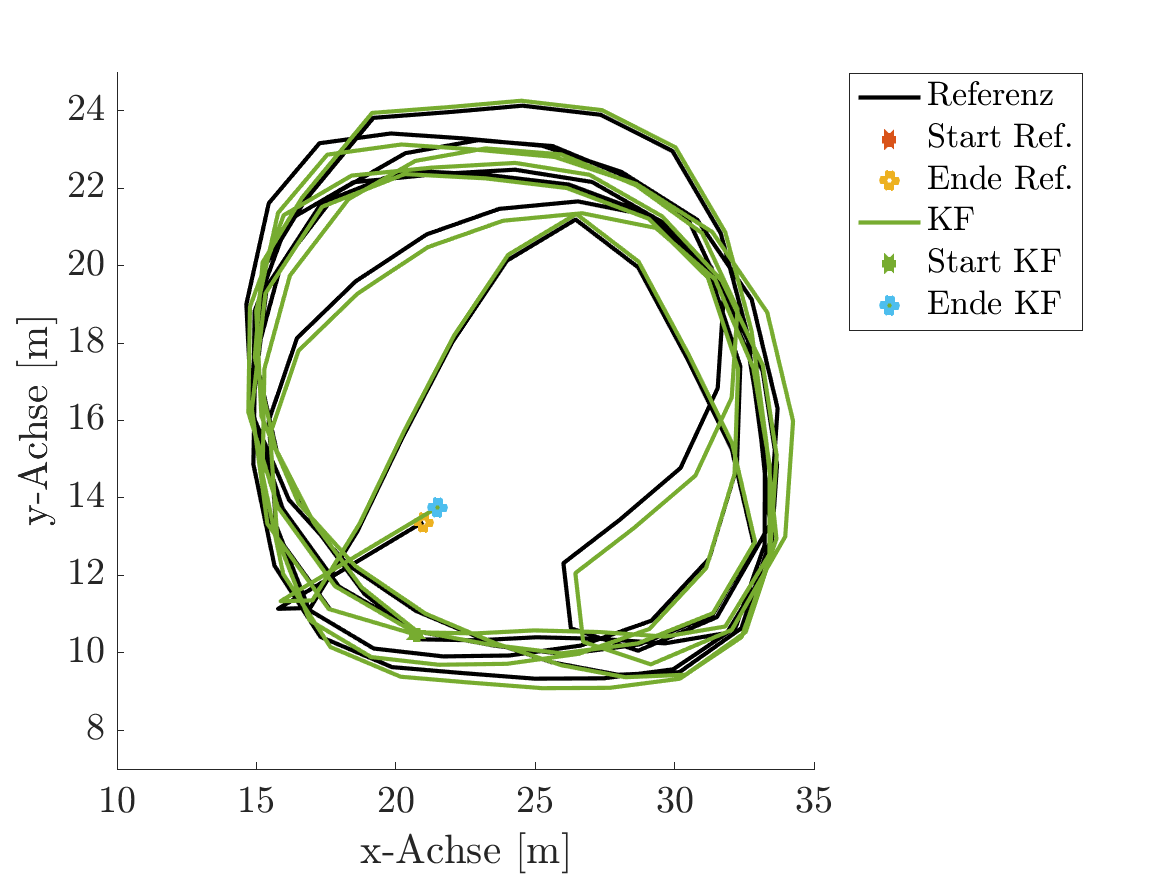}
        \subcaption{KF, \textit{jogging}.}
		\label{fig:pose:results:data:jogging_KF}
    \end{minipage}
    \hfill
	\begin{minipage}[t]{0.493\linewidth}
        \centering
    	\includegraphics[trim=30 35 180 0, clip, height=3.25cm]{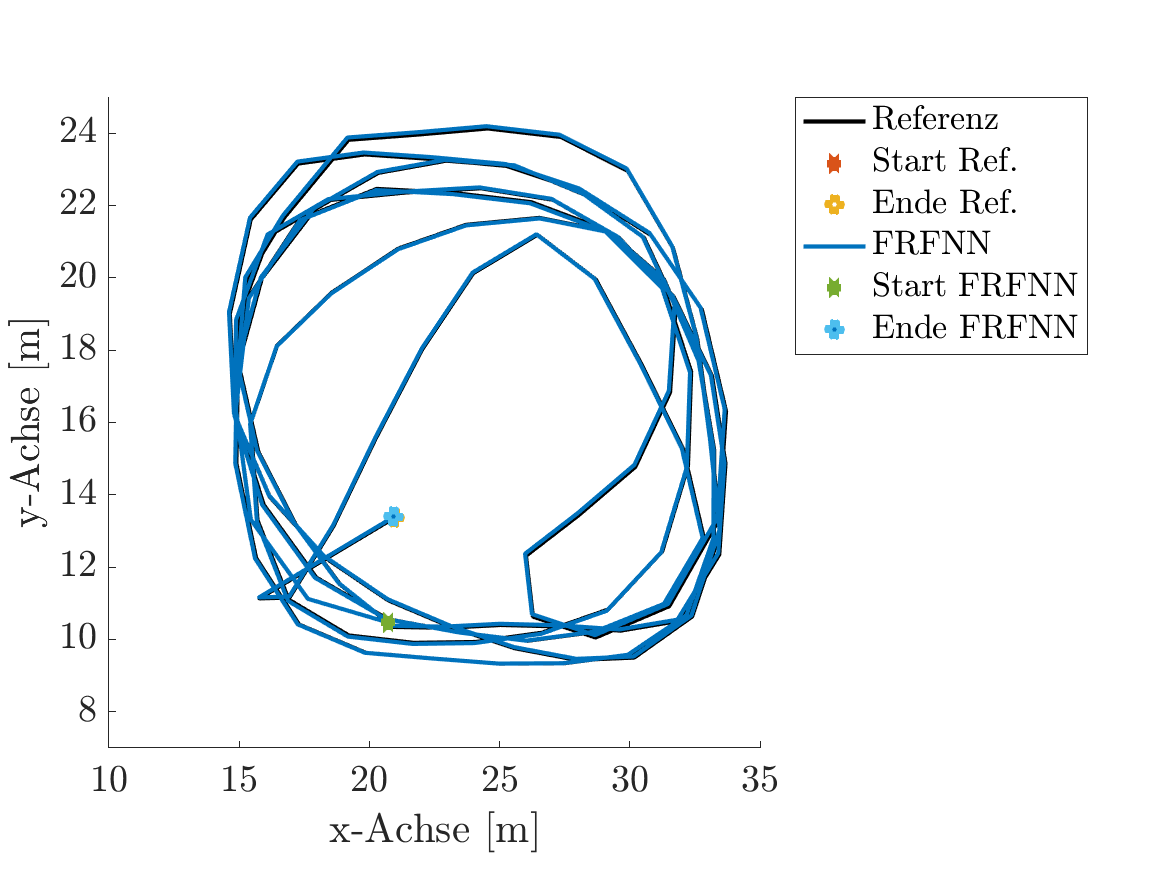}
        \subcaption{PDRNN, \textit{jogging}.}
		\label{fig:pose:results:data:jogging_PDRNN}
    \end{minipage}
    \hfill
	\begin{minipage}[t]{0.493\linewidth}
        \centering
    	\includegraphics[trim=30 25 160 0, clip, height=3.25cm]{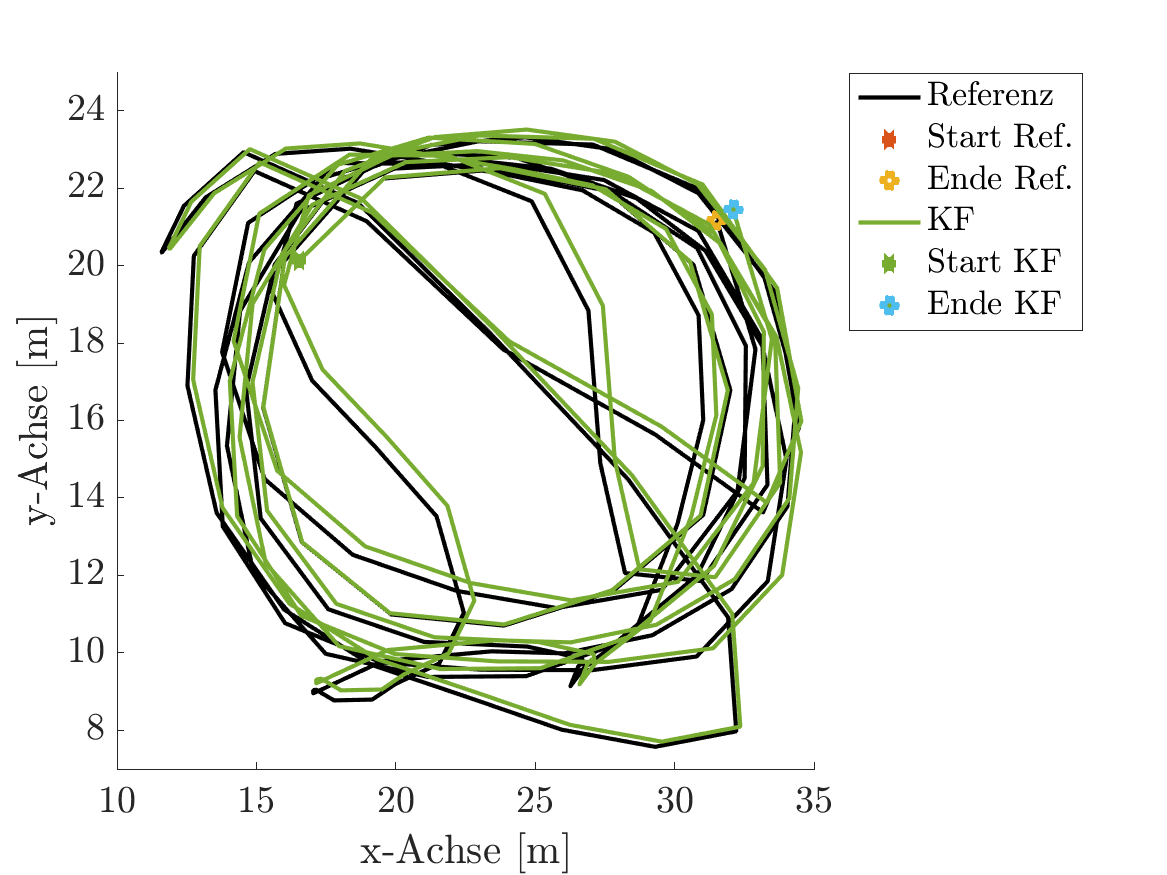}
        \subcaption{KF, \textit{running}.}
    \end{minipage}
    \hfill
	\begin{minipage}[t]{0.493\linewidth}
        \centering
    	\includegraphics[trim=30 35 180 0, clip, height=3.25cm]{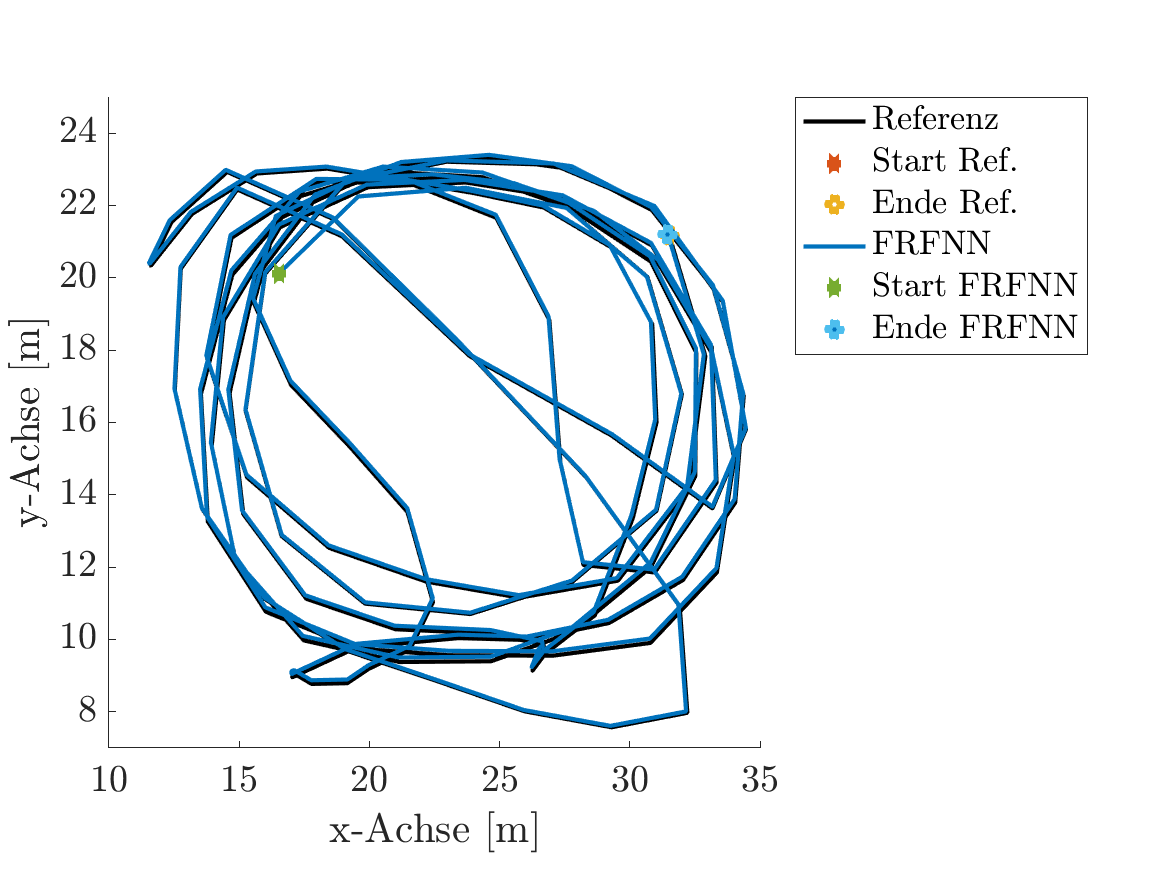}
        \subcaption{PDRNN, \textit{running}.}
		\label{fig:pose:results:data:running_PDRNN}
    \end{minipage}
    \hfill
	\begin{minipage}[t]{0.493\linewidth}
        \centering
    	\includegraphics[trim=30 25 160 0, clip, height=3.25cm]{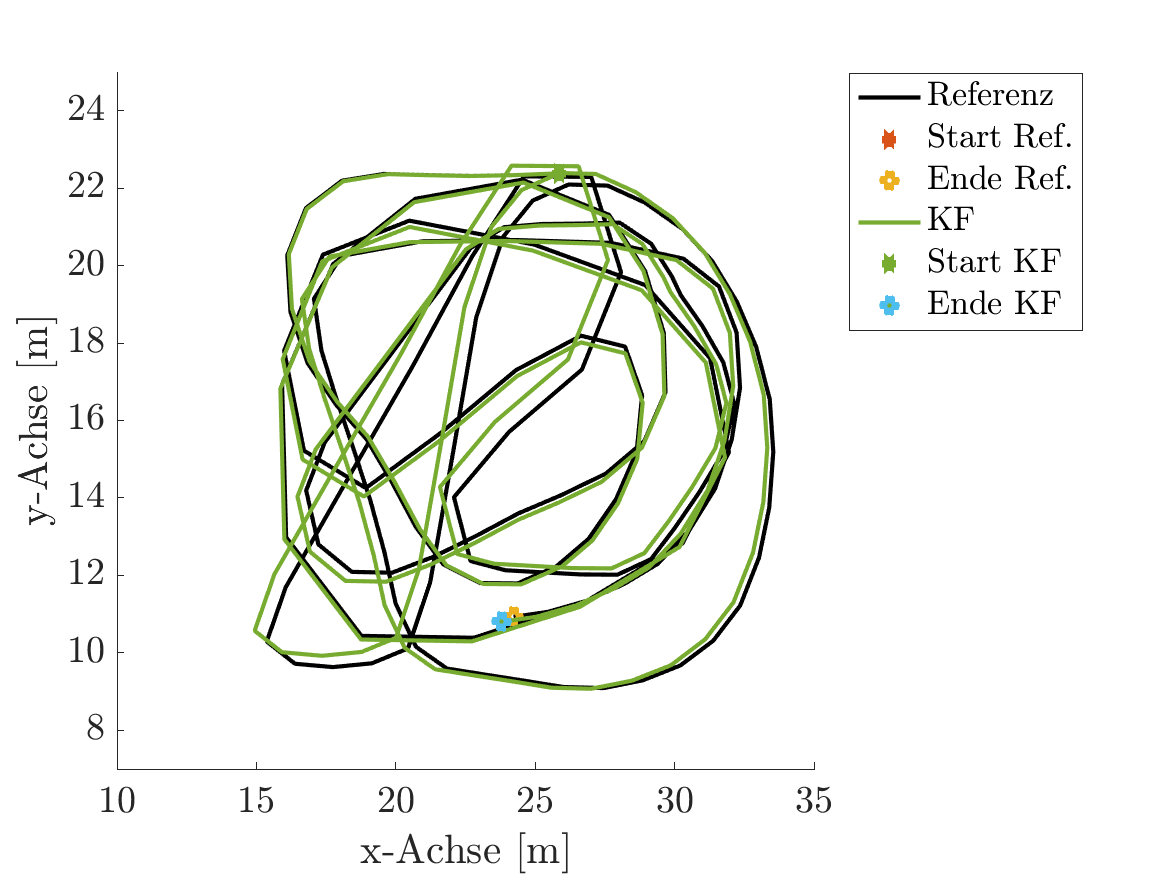}
        \subcaption{KF, \textit{random}.}
		\label{fig:pose:results:data:random_KF}
    \end{minipage}
    \hfill
	\begin{minipage}[t]{0.493\linewidth}
        \centering
    	\includegraphics[trim=30 35 180 0, clip, height=3.25cm]{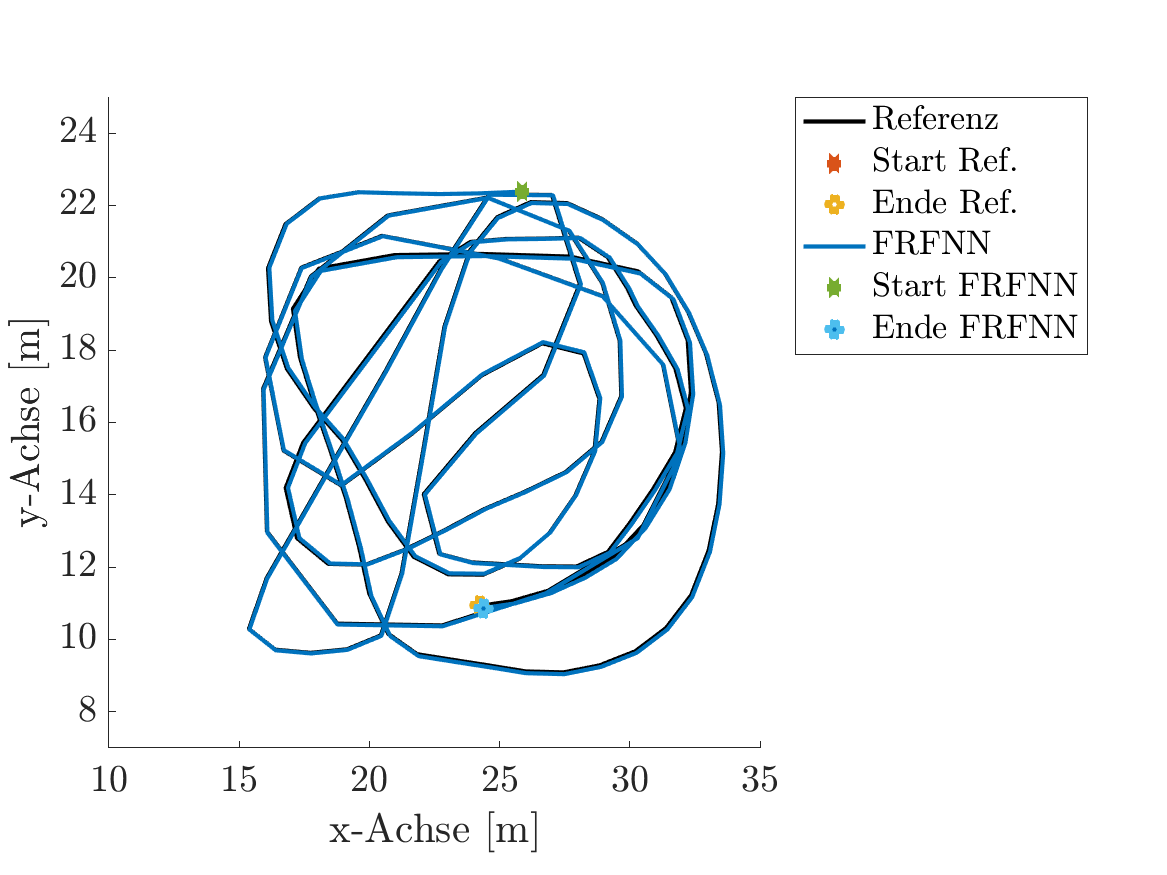}
        \subcaption{PDRNN, \textit{random}.}
		\label{fig:pose:results:data:random_PDRNN}
    \end{minipage}
    \caption{Trajectories predictions for $3\,min$ of \textit{walking}, \textit{jogging}, \textit{running} and \textit{random} of the left out test person A (x- and y-axes in $m$; black line: reference; colored line: estimates).}
	\label{fig:pose:results:data:suddenchanges-kf-PDRNN}
    \vspace{-0.5cm}
\end{figure}

\textbf{Effect of Sudden Changes in Movement.} The models' predictions for this dataset are evaluated by analyzing the MSE at each timestep in a window, providing insight into how errors evolve following sudden movement changes. This analysis also enables visualization and assessment of the transient response. As shown in Figure~\ref{fig:pose:results:data:suddensequence}, PDRNN (trained on $p_\text{radio}$, $v$) settles much faster and achieves a lower MSE of $0.00072\,m$ compared to the optimized KF ($\text{MSE} = 0.0895\,m$) even for random movements. Early experiments indicate that PDRNN maintains a low MSE ($\text{MSE} = 0.0088\,m$) and exhibits rapid convergence even when forecasting poses one second into the future. Table~\ref{table:pose:results:data:results_accuracy_trajectory_sudden} summarizes the pose accuracies of KF and PDRNN for excluded subjects. To evaluate the settling duration, ten random samples of sudden movement changes are analyzed, counting the timesteps required to return to the reference trajectory after deviations. On average, PDRNN corrects forecast errors within $0.4\,s$, while KF requires $1.8\,s$. Figure~\ref{fig:pose:results:data:suddenchanges-kf-PDRNN} visualizes the pose estimates of KF (left column, green) and PDRNN (right column, blue) on data from left-out test subject A (activities include walking, jogging, running, and random movements). While KF displays larger deviations from the reference trajectory as movement speed and directional changes increase, PDRNN consistently delivers precise and near-identical poses for all activities. Specific optimization for individual activities could potentially enhance their accuracy. The results highlight the rigidity of the KF’s predefined motion model, that is insufficient for accommodating sudden and strong variations in measurement noise.

\begin{figure}[!t]
    \centering
	\begin{minipage}[t]{0.493\linewidth}
        \centering
    	\includegraphics[trim=0 0 0 0, clip, height=3cm]{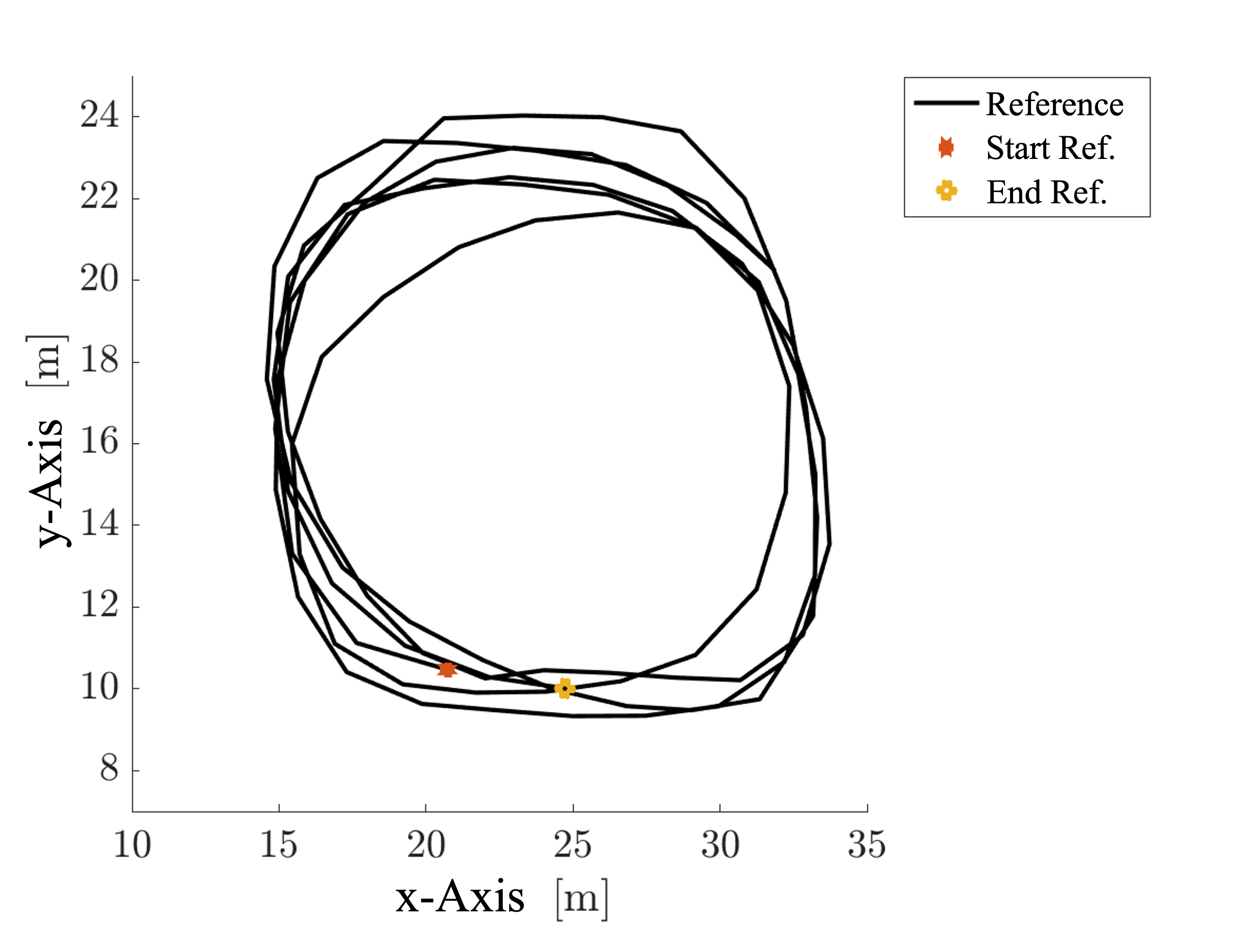}
        \subcaption{Training data.}
    	\label{fig:pose:results:data:general_ref}
    \end{minipage}
    \hfill
	\begin{minipage}[t]{0.493\linewidth}
        \centering
    	\includegraphics[trim=0 0 0 0, clip, height=3cm]{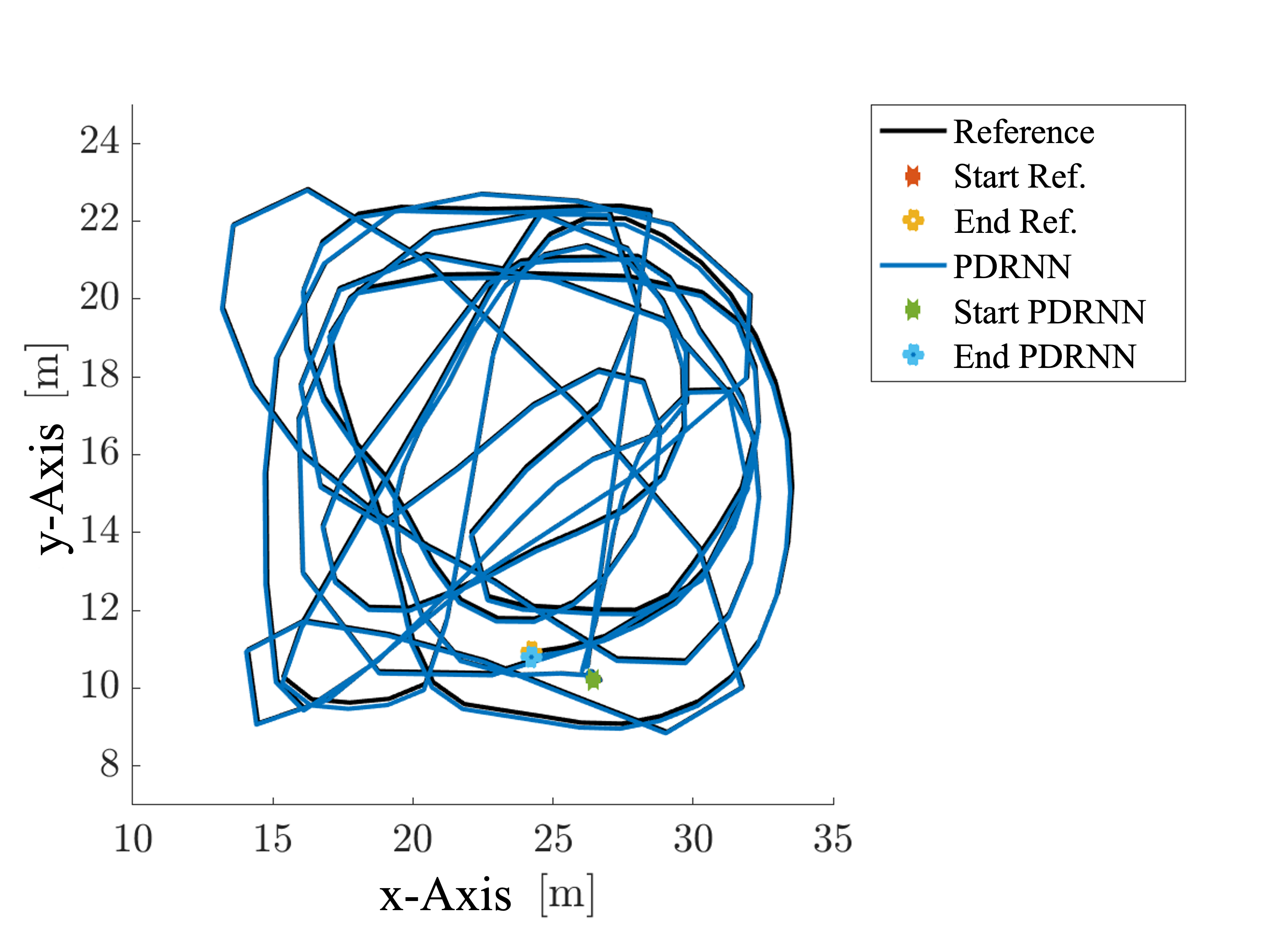}
        \subcaption{Test data.}
        \label{fig:pose:results:data:general_PDRNN}
    \end{minipage}
    \caption{Top-down view of exemplary trajectory patterns in the training set (left, approximately $3\,min$) and test set (right, approximately $3\,min$) used to assess the generalizability of PDRNN across different motion patterns. The training data, comprising activities such as \textit{walking}, \textit{jogging}, and \textit{running} from dataset V3, primarily feature circular trajectories. Instead, the test data for the \textit{random} activity from the excluded test subject A include more complex movement patterns.} \label{fig:pose:results:data:shapes}
    \vspace{-0.5cm}
\end{figure}

\textbf{Effect of Unknown Trajectory Shapes.} The KF handles trajectory estimation well, aside from occasional over- and undershooting, as its motion model is based on the physical states $p_\text{radio}$, $v$, and $acc$, rather than specific motion shapes. Consequently, KF achieves a positional accuracy comparable to models explicitly optimized for random activities, extracting physical relationships directly from the data ($\text{MAE} = 0.4427$, $\text{MSE} = 0.2580$, $\text{RMSE} = 0.5001$, $\text{CEP}_{95} = 0.8331\,m$). Instead, when PDRNN is trained solely on circular motions during \text{walking}, \textit{jogging}, and \textit{running}, it produces implausible results for random activities. This failure arises as central positions within circular trajectories remain unknown, and many essential basic motion patterns are missing, see Figure~\ref{fig:pose:results:data:general_ref}. These patterns are critical for reconstructing complex motions. To address this limitation, PDRNN was retrained using all data from the three activities in dataset V3, that includes paths where athletes reverse directions or cross circular trajectories. This improved model accurately estimates poses for random activities of excluded subjects, as illustrated in Figure~\ref{fig:pose:results:data:general_PDRNN}. These results demonstrate that training data must cover the fundamental building blocks of motion patterns to enable accurate reconstruction of unknown complex movements. Shorter sequence lengths of 128 values yielded the best results ($\text{MAE} = 0.0417$, $\text{MSE} = 0.0041$, $\text{RMSE} = 0.0821$, $\text{CEP}_{95} = 0.0612\,m$). Conversely, longer sequences of 256 values significantly degraded accuracy ($\text{MAE} = 0.3675$, $\text{MSE} = 0.2168$, $\text{RMSE} = 0.3273$, $\text{CEP}_{95} = 0.6578\,m$). The poorer performance with longer sequences may stem from their inability to capture diverse random movements, that shorter sequences can better represent by incorporating curvy and straight components. Longer sequences, that often model entire circular paths, lack the flexibility to generalize to arbitrary trajectory changes. Figure~\ref{fig:pose:results:data:general_PDRNN} depicts a reconstructed trajectory comprising numerous random movements. PDRNN reconstructs highly congruent and precise trajectories, even when trained on less complex motion forms, see Figure~\ref{fig:pose:results:data:suddenchanges-kf-PDRNN}.

\textbf{Impact of Data Gaps.} At least two radio positions are necessary to interpolate missing positions using velocities. With a sequence length of $1.28\,s$, the position update rate can be reduced from 100\,Hz to 1\,Hz, thereby decreasing the communication load by a factor of 100. Initial experiments demonstrate that incorporating a single radio position per input sequence or window can yield accurate pose estimates if the directed distance (delta) between successive positions is integrated into the input rather than the absolute position. In this approach, instead of predicting an explicit position, the model predicts the directed distance to the next position. The estimated distance is then added to a starting position or each newly estimated position. This technique achieved comparable accuracy to the method utilizing two radio positions per window, yielding results such as $\text{MAE} = 0.0686$, $\text{MSE} = 0.0034$, $\text{RMSE} = 0.0066$, and $\text{CEP}_{95} = 0.1546\,m$.
\section{Conclusion}
\label{sec:summary}

This paper introduces a significant advancement in the forecasting of human movement through a novel data-driven method, PDRNN, which effectively integrates radio frequency and inertial sensor data in a PDR framework, utilizing modular ML components at each stage. The architecture uniquely combines forward-coupled layers with LSTM cells, enabling the extraction of spatio-temporal features from sensor signals and their translation into meaningful motion data. The results demonstrate the limitations of traditional approaches, such as KF and model-based PDR methods, particularly in dynamic environments characterized by abrupt motion changes. PDRNN outperforms monolithic ML-based PDR techniques, such as RONIN, offering superior accuracy and efficiency. It effectively handles sensor noise and data gaps, providing reliable pose and trajectory estimations. 
PDRNN exhibits the ability to generalize across a range of activities and user behaviors, adapting seamlessly to diverse trajectories and velocities without significant loss of accuracy. Key innovations include the model's ability to generalize across various activities and user behaviors, enabling it to adapt to different movement patterns while maintaining high accuracy. The model's robust handling of sensor noise and data gaps ensures reliable performance under challenging conditions. Exploiting positions and velocities as input enhances overall accuracy, improves real-time processing capabilities, and facilitates the prediction of future poses, enabling compensation for system delays. 

Our experiments underscore the potential of PDRNN for precise pose estimation in dynamic environments. The model addresses key challenges through a modular architecture that integrates prominent ML modules (RoNIN, LSTM) to effectively capture temporal relationships within sensor data while providing uncertainty measures. The results demonstrate that PDRNN learns precise, adaptable, and generalizable motion patterns, outperforming established methods, particularly during sudden or random movement changes ($\text{CEP}_{95}$: $\text{PDR} = 1.25\,m$, $\text{RoNIN} = 0.46\,m$, $\text{PDRNN} = 0.14\,m$). It exhibits shorter settling times and lower overall error rates under these conditions. Notably, PDRNN predicts random trajectories up to one second into the future with errors below $0.12\,m$. Interestingly, increasing network depth did not yield significant accuracy gains, indicating that deeper layers were not critical to the model's performance. This highlights the efficiency and practicality of the proposed approach in managing complex human motion while maintaining high prediction and pose estimation accuracy close to the sensors.

\bibliography{ION_PLANS}
\bibliographystyle{IEEEtran}

\end{document}